\documentclass[journal]{IEEEtranTIE}
\usepackage{graphicx}
\usepackage{cite}
\usepackage{picinpar}
\usepackage{amsmath}
\usepackage{url}
\usepackage{flushend}
\usepackage{colortbl}
\usepackage{soul}
\usepackage{multirow}
\usepackage{pifont}
\usepackage{color}
\usepackage{alltt}
\usepackage[hidelinks]{hyperref}
\usepackage{enumerate}
\usepackage{siunitx}
\usepackage{breakurl}
\usepackage{epstopdf}
\usepackage{pbox}
\usepackage{amsmath,amssymb,amsfonts}
\usepackage{subfigure}
\usepackage{cite}
\usepackage{balance}
\usepackage{flushend}

\newcommand{\diag}{{\rm diag\;}}
\newcommand{\dfb}{\stackrel{\Delta}{=}}

\newtheorem{remark}{Remark}

\newtheorem{theorem}{Theorem}

\begin{document}
\title{Self-organized Polygon Formation Control based on Distributed Estimation}

\author{
	\vskip 1em
	
	Qingkai Yang, \IEEEmembership{Member, IEEE}, Fan Xiao, Jingshuo Lyu, Bo Zhou and Hao Fang, \IEEEmembership{Member, IEEE}

	\thanks{This work was supported in part by the National Key Research and Development Program of China under No. 2022YFA1004703, No. 2022YFB4702000, in part by the NSFC under Grants 62133002, U1913602, 62088101, 61720106011 and in part by the Shanghai Municipal Science and Technology Major Project (2021SHZDZX0100)..} 
	\thanks{Corresponding author:Hao Fang}
	\thanks{All the authors are with the School of Automation, Beijing Institute of Technology, Beijing 100081, China.
		   {\tt\small (email: fangh@bit.edu.cn) }}%
}

\maketitle
	
\begin{abstract}
	This paper studies the problem of controlling a multi-robot system 
	to achieve a polygon formation in a self-organized manner.
	Different from the typical formation control strategies 
	where robots are steered to satisfy the predefined control variables, 
	such as pair-wise distances, 
	relative positions and bearings, 
	the foremost idea of this paper is to achieve polygon formations 
	by injecting control inputs randomly to a few robots (say, vertex robots) of the group, 
	and the rest follow the simple principles of moving towards the 
	midpoint of their two nearest neighbors in the ring graph without any external inputs. 
	In our problem, a fleet of robots is initially distributed in the plane. The so-called vertex robots take the responsibility of determining the geometric shape of the entire formation and its overall size, while 
	the others move so as to minimize the differences with two direct neighbors.
	In the first step,
	each vertex robot estimates the 
	number of robots in its associated chain. 
	Two types of control inputs that serve for the estimation are designed using the measurements from the latest
	and the last two time instants respectively.
	In the second step,
	the self-organized formation control law is proposed 
	where only vertex robots receive external information.
	Comparisons between the two estimation strategies are carried out in terms of the convergence speed and robustness. The effectiveness of the whole control framework is further validated in both simulation and physical experiments.
\end{abstract}


\markboth{IEEE TRANSACTIONS ON INDUSTRIAL ELECTRONICS}%
{}

\definecolor{limegreen}{rgb}{0.2, 0.8, 0.2}
\definecolor{forestgreen}{rgb}{0.13, 0.55, 0.13}
\definecolor{greenhtml}{rgb}{0.0, 0.5, 0.0}

\section{Introduction}

\IEEEPARstart{M}{ulti-robot} systems have attracted intensive attention in recent years.
In general, the robots cooperate with each other to overcome 
the shortcomings of limited computational resources and local communication/sensing capabilities.
The cooperative control of multi-robot systems is broadly used in search and rescue \cite{13}, 
 transportation and construction \cite{transport}, 
mapping and navigation \cite{16}, sensor network deployment \cite{17}, etc. 

The primary goal of formation control is to drive a multi-robot system to form the prescribed geometric shape, 
which serves as an important module for complex tasks.
In typical consensus-based formation control strategies \cite{2007Consensus,consensus2,zhang_consensus_2022,angle}, robots are driven to achieve the desired control variables such as relative position, distance, bearing and angle, the values of which are consistent with the prescribed formation, and thus the convergence of control variables results in the realization of formation control.   
{\color{black}To make the swarm more autonomy and adapted, some recent research attempt to use less priori calibrated information during formation.}
The complex Laplacian employed in \cite{complex} can reduce the number of informed agents.
In \cite{complexmaneuver,complexscale}, it is shown that transformations including scaling, rotation and translation can be realized by only controlling the leaders.
As an extension, a matrix-valued Laplacian is introduced to gain more flexibility in dynamic formation change \cite{matrixweight}. 
In \cite{necessary}, the information of the desired formation is encoded into the stress matrix, enabling the convergence to its affine image by only controlling three leaders.
Moreover, as an alternative way to relieve the dependence on the exact knowledge of formation parameters, some estimation methods are developed to infer the system states \cite{size}, formation scaling size \cite{sizeestimate,Yang}, and mixed scaling and rotation variables \cite{yang2}, to name a few.      
{\color{black}However, it is required in most of the existing methods that all the desired pair-wise control variables have to be pre-defined carefully before its implementation, which is of huge computation complexity. The tedious pre-defined procedure also reduces the feasibility to the changing tasks or the ambient environment.}

It has been observed that the collective behavior of swarms in nature are almost self-organized,
such as the aggregation of birds and fish, and the social structure of ant colony,
that is, via very simple interaction principle among neighbors, the swarms can form different patterns to adapt to environment changes. 
Motivated by this fact, by introducing the concept of morphology into swarms,
self-organized rules and emergence behaviors are exploited on simple mobile robots to
obtain a variety of spatial configurations \cite{57}.
To verify the capability of creating emergent morphologies via purely self-organizing behaviors, 300 simple robots are put into use without any self-localization \cite{morpho}.
Recently, it has been proved that less communication can contribute to better adaptation to changes by using the specified voter model \cite{lessismore}. 
Besides, from the perspective of micro-world, 
gene regulatory network is utilized in \cite{ICIRA,66}, where each robot contains two genes generating proteins to control the movement of robots.
It is also reported in \cite{IKEMOTO} that a group of robots can gradually generate some complicated patterns such as a polygon by using the Turing diffusion-driven instability theory
where two signals exchange between the swarms through a set of reaction–diffusion differential equations.
{\color{black}However, these self-organized methods make it challenging to form a specified desired shape.}

\begin{figure}
	\centering
	\includegraphics [width=0.5\textwidth]{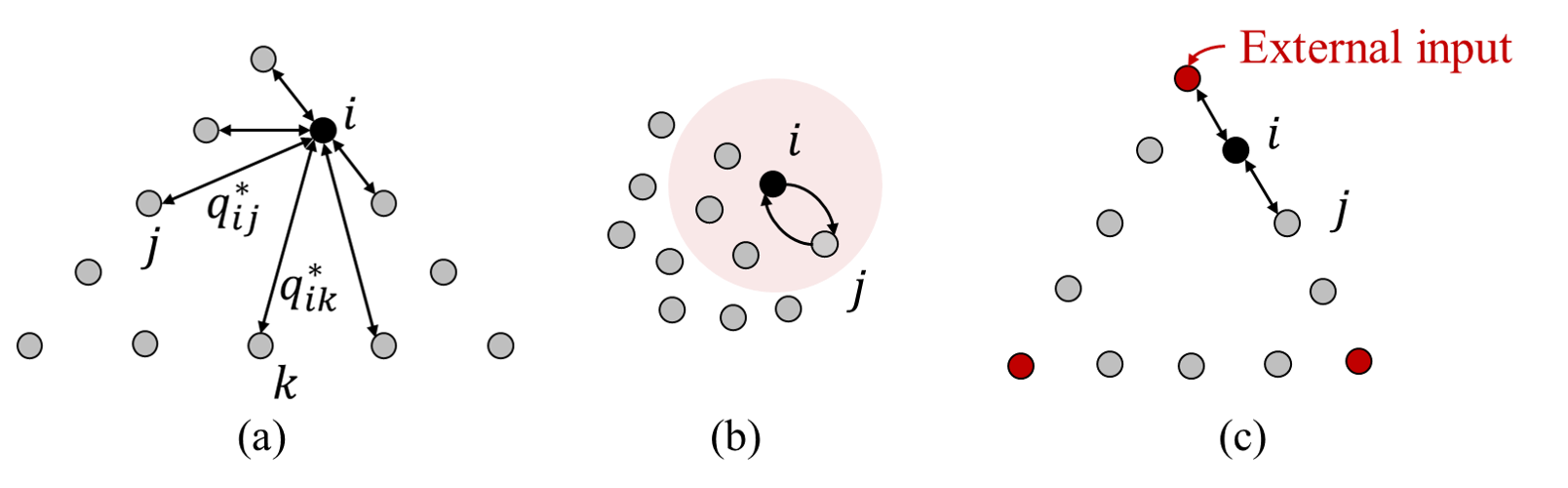}
	\caption{Intuitive comparison from the perspective of mutual interaction. (a) the consensus-based formation control \cite{2007Consensus} $\dot{p_{ij}}=\Sigma a_{ij}(p_i-p_j-p_{ij}^*)$, which requires the pre-defined desired relative positions among all neighbors; 
	(b) the purely self-organizing morphogenesis \cite{morpho} $\dot{q_i}=Rf(q_i,q_j)+D_{q_i}\nabla ^2q_i,\dot{q_j}=Rg(q_i,q_j)+D_{q_j}\nabla ^2q_j$, where robots in the red shadow zone are considered as the neighbors of $i$; 
	(c) the proposed method, where the external interventions are exerted on the vertex robots (generally a few among the group), and the others only interact with its two direct neighbors.}\label{compare}
\end{figure}

This paper focuses on the problem of self-organized deterministic polygon formation control for swarm robots with the aid of a few external interventions exerted on the vertex robots. 
The sensing topology among the robots is cyclic, where each robot can only interact with its two direct neighbors.
To make the problem tractable, we first divide the whole ring topology into virtual segments, and each vertex robot estimate the number of robots in its associated chain only using local measurements. 
Then, with the accurate estimation value, 
the vertex robots actively move to adjust the collective formation shape as well as its scaling size, playing the role of shepherd dogs when herding sheep. 
The others move according to their intrinsic interaction with their two direct neighbors. 
The intuitive comparison of the above-mentioned methods can be seen in Fig. \ref{compare}. 
In the consensus-based formation control framework, 
the desired relative positions among all neighbors have to be pre-defined carefully. 
In the strategy of purely self-organizing morphogenesis, each robot only interacts with its neighbors without any external injection to generate emergence behaviors, 
while it is generally hard to obtain prescribed formation patterns.
In contrast, the proposed control strategy can achieve any specific polygon formation 
determined by the relative positions of vertex robots instead of all pairwise robots, 
which significantly reduces the computation complexity and the number of control variables. 
Another distinguishing feature of the proposed scheme is the scalability in the sense that 
the desired formation is defined by a few among many, 
which allows flexible joining and leaving without altering the stabilized formation shape. 

The rest of this paper is organized as follows.
In Section \ref{PRELIMINARIES}, some notations and preliminary theories are given, as well as the problem to be addressed. 
Two classes of distributed controllers for estimation are proposed in Section \ref{ESTIMATION} to derive the cardinality of the associated robot set.
Then the formation control strategies are designed in Section \ref{formation}.
Finally, the simulation and experimental results are presented in Section 
\ref{simulation}, followed by the conclusion in Section \ref{conclusion}.

\section{PRELIMINARIES}\label{PRELIMINARIES}
This section will give basic knowledge of notations, the related graph theory and the statement of the problem to be addressed. 
\subsection{Notations}
Let $\mathbb{R}^{n\times m}$, $\mathbb{R}^n$, and $\mathbb{R}$ denote the sets of 
real matrices (of dimension $n\times m$), real vectors (of dimension $n$) and real numbers, respectively.
Let $\mathbf{0}$ be the matrix with all entries equal to zero
and $I$ be the identity matrix.
The symbol $|\cdot |$ represents the absolute value of a real number, the magnitude of a complex number, and the determinant of a matrix, respectively. 
we use $\| x\|$ to denote the $2$-norm of a vector $x$. 
Given two sets $A$ and $B$, the subtraction operation is indicated by $A-B$, i.e., removing the elements belong to the set $B$ from $A$.   

\subsection{Graph theory}
In this paper, 
the interaction among the networked robots 
is described by an undirected graph $\mathcal{G} =(\mathcal{V}, \mathcal{E}, \mathcal{A} )$,
where $\mathcal{V} =\{0,1,\dots,n-1\}$ is the node set,
$\mathcal{E} \in \mathcal{V} \times\mathcal{V} $ is the edge set
and $\mathcal{A} =[a_{ij}]\in \mathbb{R} ^{n\times n}$
represents the binary adjacency matrix with $a_{ij}=0$ if $(i,j)\notin \mathcal{E} $
and $a_{ij}=1$ otherwise.
The neighbor set is defined as $\mathcal{N} _i=\{j|(i,j)\in \mathcal{E} \}$.
The edge $(i,j)\in \mathcal{E} $ indicates
that robots $i$ and $j$ can sense each other.
Now we introduce two kinds of undirected graphs. 

\begin{itemize}
    \item \emph{Ring graph}: a cyclic graph where the neighbors of node $i$ are nodes $i-1$ and $i+1$ (mod $n$)\cite{Cyclic}.
    \item \emph{Chain graph}: an connected graph that all the nodes have two neighbors except for two ending nodes who have only one neighbor.
\end{itemize}

\subsection{Polygon formation}
A \emph{configuration} $q\in \mathbb{R} ^{n\times 2}$ is a finite collection of the positions of $n$ labeled robots, denoted by $q=[q_0,q_1,\dots q_{n-1}]^T$.
A framework $(\mathcal{G} ,q)$ is obtained by assigning a feasible configuration $q$ 
to its associated graph $\mathcal{G}$ in the Euclidean space.
In a polygon formation,
a robot is called the \emph{vertex robot} if it is non-collinear with its neighbors.
Assume that the abstracted polygon has $m$ vertices, and the corresponding vertex robots are collected in the set $S=\{s_0,s_1,\cdots,s_{m-1}\} \subset \mathcal V$. Note that the non-negative integers $s_i$ and $s_{i+1}$ are not necessarily consecutive. 
For vertex robots  $s_i$ and $s_{i+1}$, we define  $n^s_i =s_{i+1}-s_i$ as the number of their in-between nodes. 
The stacked form is given by $n^s=[n^s_0,n^s_1,\dots,n^s_{m-1}]^T$. Correspondingly, the relative positions between vertex robots are concatenated in the vector $r=[r_0,r_1,\dots,r_{m-1}]^T$ with $r_i=q_{s_i}-q_{s_{i+1}}$. 
{\color{black}In this paper, the number of vertex robots needs to be consistent with the number of vertices. However, the vertex robot is label-free, which means the index in the set S may change as robots move. 
This contributes to the scalability of the swarm and the flexible change of the desired polygon formation, which stimulates the self-organized collective behavior.}

\subsection{Problem formulation}\label{PROBLEM STATEMENT}
\begin{figure}
	\centering
	\subfigure [The ring graph $\mathcal{G}$]{
		\includegraphics [width=0.15\textwidth]{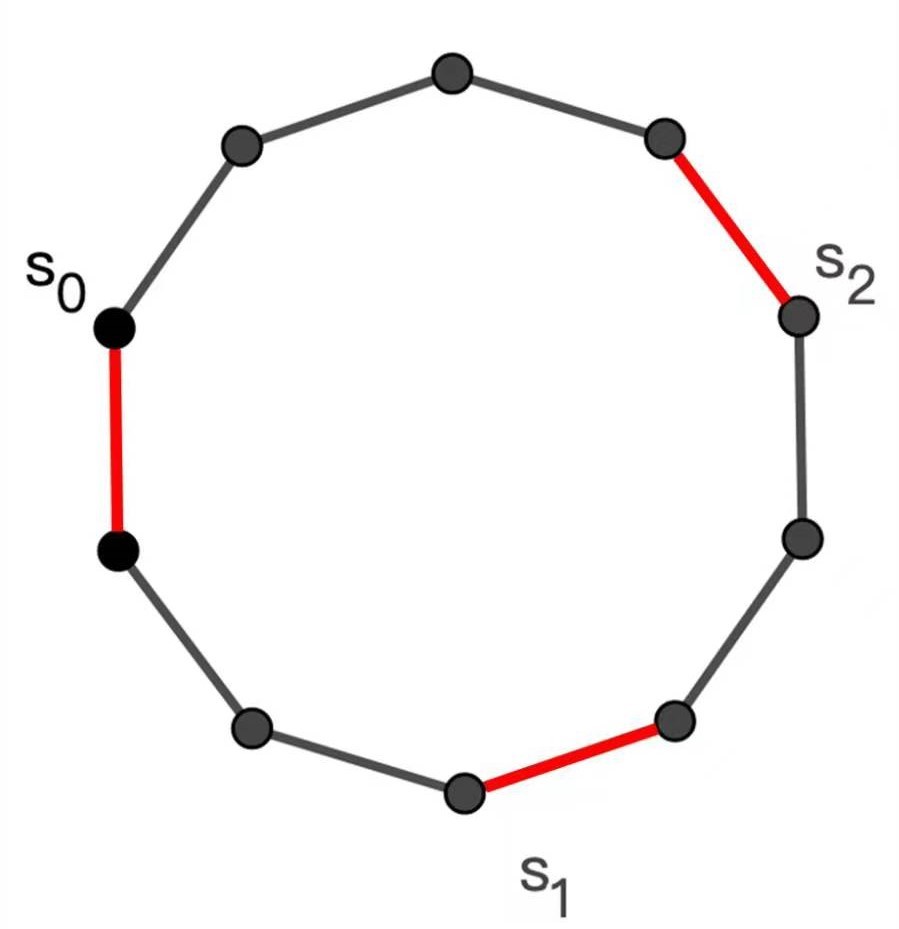}}
	\hspace{.2cm}
	\subfigure [The resultant subgraph $\mathcal{G}_e$]{
		\includegraphics [width=0.15\textwidth]{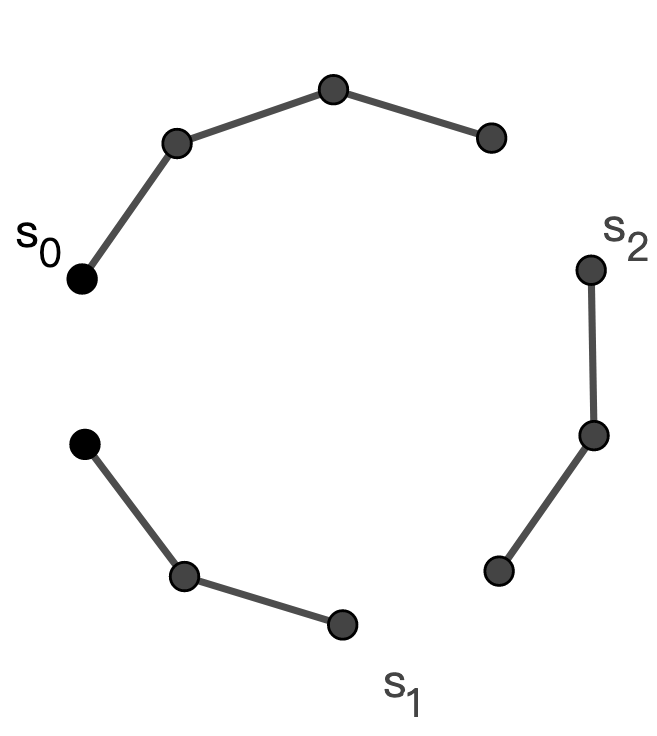}}
	\caption{The ring graph and the resultant chains after cutting operation. 
		  }
    \label{ring}
\end{figure}

This paper focuses on the formation control of $n$ robots modelled by 
discrete-time dynamics
\begin{equation}\label{sys:model}
	\begin{aligned}
		&q_i(k+1) = q_i(k)+\Delta t * v_i(k),
	\end{aligned}  
\end{equation}
where $q_i\in \mathbb{R}^2$ represents the position of robot $i$
and $\Delta t$ denotes the time interval between two sampling instants.
The robot team is expected to form a polygon shape with $m$ vertices.
The only injected information for the robot team is the desired relative position
between vertex robots, i.e., $r^*=[r_0^*,r_1^*,\dots,r_{m-1}^*]^T$, whose component $r_i^*$ is only available to vertex robot $s_i$. Except for {\color{black}such} `external information', all the robots are self-regulated via local sensing and communication. 
The cardinality of the set $\mathcal V$, i.e., the integer $n$, and the number of robots along each edge of the polygon are unknown. The communication/sensing relationship is represented by the ring graph. It can be seen from Fig. \ref{ring} that after removing the red edges incident to vertex robots, say \emph{cutting operation}, we obtain the subgraph $\mathcal{G}_e$ composed of $m$ chains.

Aiming to present a comprehensive and trackable solution, we decompose the overall self-organized polygon formation control problem into two sub-problems. 
First, the distributed estimation problem conducted by vertex robot $s_i$ to infer the number of robots along the chain where it stays, i.e., $s_{i}-s_{i-1}$.  
Then, the control objective is to design the distributed law for each
robot $i$ using only local information to achieve the desired polygon formation, which is represented by $r_i^*$ even though it is unknown to most of the robots.

\section{DISTRIBUTED ESTIMATION}
\label{ESTIMATION}
Without loss of generality, we consider the estimation problem along one specific chain with $n$ robots, grouped in the set $\mathcal V_C$ , and thus the neighbor sets are given by $\mathcal{N} _0=\{1\},\mathcal{N} _i=\{i-1,i+1\},i\in \mathcal{V}-\{0,n-1\},\mathcal{N} _{n-1}=\{n-2\}$.
The robot $n-1$ needs to estimate the unknown integer $n$. Two strategies utilizing different historical data are proposed, and the rigorous theoretical analyses are also given. 

\subsection{Estimation based on the latest measurements}
\label{sec:3a}
In this subsection, assume that only the measurements from the latest sampling instant are available.
The distributed controller for activating the estimation process is designed as
\begin{equation} \label{estimator}
	\begin{aligned} 
		v_{0}(k+1)=&0 \\
		v_{i}(k+1)=&\frac{\alpha}{2}\left(q_{i+1}(k)+q_{i-1}(k)-2 q_{i}(k)\right)\\&+\frac{v_{i+1}(k)+v_{i-1}(k)}{2}, i \in \mathcal{V}_C-\{0\} \\
		v_{n}(k+1)=&-v_{n}(k) \\
	\end{aligned}
\end{equation}
where $\alpha$ is a positive constant.
Assume that robot $0$ stays still at the origin all the time.
It is worth noting that robot $n$ is a virtual one, 
which means $v_n(k)$ can be regarded as an excitation signal.
Under the controller (\ref{estimator}),
robots in the chain will act like a stable oscillator when the convergence is reached.
Let $n'\dfb n-1$ for simplicity.
Recall that the $n'$th robot needs to estimate the total number of robots moving in the chain, namely the value of $n'+1$.
Instead of directly estimating $n^\prime +1$, we seek to figure out the value of $n^\prime$  
using the states of $n'$th robot.

{\color{black}\begin{remark}\label{rem:remark}
	The controller can be transformed into
	\begin{equation}
		\begin{aligned} 
			&(v_{i+1}(k)-v_{i}(k+1)) - (v_{i}(k+1)-v_{i-1}(k))\\&=\alpha [(q_{i+1}(k)-q_{i}(k))-(q_{i}(k)-q_{i-1}(k))], i \in \mathcal{V}_C-\{0\},\\
		\end{aligned}
	\end{equation}
	The relative position and velocity of robot $j$ measured in robot $i$'s local reference frame can be expressed as
	$p_{ij}^{(i)}=R_ip_{ij}$, $v_{ij}^{(i)}=R_iv_{ij}$, where $R_i$ is the rotation transformation from the global frame to the local frame of robot $i$.
	Then, the control law can be written as
	$R_i^{-1}(v_{i+1}^{(i)}(k)-v_{i}^{(i)}(k+1)) - R_i^{-1}(v_{i}^{(i)}(k+1)-v_{i-1}^{(i)}(k))=\alpha [R_i^{-1}(q_{i+1}^{(i)}(k)-q_{i}^{(i)}(k))-R_i^{-1}(q_{i}^{(i)}(k)-q_{i-1}^{(i)}(k))]$.
	Multiplying the above equation by the rotation matrix $R_i$ from the left side,
	the controller expressed in the local coordinate frame is obtained as
	$v_{i+1}^{(i)}(k+1) = \alpha /2 [(q^{(i)}_{i+1}(k)-q^{(i)}_{i}(k))-(q^{(i)}_{i}(k)-q^{(i)}_{i-1}(k))] + (v^{(i)}_{i+1}(k)+v^{(i)}_{i-1}(k))/2$,
	which is the same as \eqref{rem:remark}.
	The relative position and relative velocity can be measured by onboard sensor, 
	but $v_j^{(i)}$ is technically difficult to measure directly in the local coordinate frame.
	Normally, $v_j^{(i)}$ is calculated by subtracting the measured relative velocity $v_{ij}^{(i)}$ from the robot $i$'s own velocity $v_i^{(i)}$.
	Moreover, in a typical application scenario where communication is allowed and the orientations of each local coordinate frame are aligned, 
	the neighbors can transmit their own velocities $v_{i+1}(k)$ and $v_{i-1}(k)$ directly to robot $i$.
\end{remark}}

Prior to giving main result on the convergence of the closed-loop system under \eqref{estimator}, we introduce an auxiliary variable $s \in \mathbb R^{2n'\times 2}$ defined by $s(k)=[q_1(k),q_2(k),\ldots,q_{n'}(k),v_1(k),v_2(k),\ldots,v_{n'}(k)]^T$, whose dynamics  satisfy 
\begin{equation}\label{s:dynamics}
	\begin{aligned}
		s(k+1)=\underbrace{\begin{bmatrix}
			I & \Delta t * I\\
			\alpha A_{21} & A_{22}    
        \end{bmatrix}}_{\dfb A} s(k)
		+bv_{n'+1}^T(k)
	\end{aligned}
\end{equation}
where $A_{21},A_{22}\in\mathbb{R}^{n'\times n'}$ are given by 
\begin{equation*}
	\small
	\setlength{\arraycolsep}{1.5pt}
	A_{21}=\begin{bmatrix}
		-1 & 0.5 & 0 & & 0 \\
		0.5 & -1 & 0.5 & \cdots & 0 \\
		0 & 0.5 & -1 & & 0 \\
		& \vdots & & \ddots & \vdots \\
		0 & 0 & 0 & \cdots & -1
	\end{bmatrix},
	\setlength{\arraycolsep}{1.2pt}
    A_{22}=\begin{bmatrix}
		0 & 0.5 & 0 & & 0 \\
		0.5 & 0 & 0.5 & \cdots & 0 \\
		0 & 0.5 & 0 & & 0 \\
		& \vdots & & \ddots & \vdots \\
		0 & 0 & 0 & \cdots & 0
	\end{bmatrix}
\end{equation*}
and $b= [0,0,\cdots, 0.5]^T \in\mathbb{R}^{2n'}$. 
By applying iterative process, \eqref{s:dynamics} turns to be  
    \begin{equation}\label{s:iteration}
        \begin{aligned}
            s(k+1)=&A^2s(k-1) + Abv_{n'+1}^T(k-1) +bv_{n'+1}^T(k)\\
            =&A^ks(1) + \sum_{i=1}^k\left(A^{k-i}bv_{n'+1}^T(i)\right).
        \end{aligned}
    \end{equation} 
It can be obtained from \eqref{estimator} that
$v_{n'+1}^T(i)= (-1)^{i-1}v_{n'+1}^T(1)$.
Substituting this equality into \eqref{s:iteration} yields
\begin{equation*}\label{3.7}
	\begin{aligned}
		s(k+1) = &A^ks(1) + \sum_{i=1}^k\left(A^{k-i}b (-1)^{i-1}v_{n'+1}^T(1)\right)\\
		= &A^ks(1)  +(-1)^{k+1}\sum_{i=1}^k\left((-A)^{k-i}bv_{n'+1}^T(1)\right).
	\end{aligned}
\end{equation*}

\begin{theorem}\label{lem3.1}
    The spectral radius of matrix $A$ is less than 1 if the parameter $\alpha$ is chosen satisfying
    $\alpha \Delta t < \frac{1-\cos^2(\frac{\pi}{n'+1})}{3-\cos^2(\frac{\pi}{n'+1})}$.
\end{theorem}
The proof of Theorem \ref{lem3.1} is given in Appendix \ref{app1}.


Under Theorem \ref{lem3.1}, there holds $\lim_{k\rightarrow \infty} A^k= 0$ and the matrix power series $\sum_k A^k$ converges. In addition, we know
$\lim_{k\rightarrow \infty}\sum_{i=1}^{k} (-A)^{k-i}=\lim_{k\rightarrow \infty}\sum_{i=0}^{k-1} (-A)^{i}=[I-(-A)]^{-1} = (I+A)^{-1}$ \cite{matrixanalysis}.
Hence, it yields   
\begin{equation}\label{eq:sk:infty}
	\begin{aligned}
		\lim_{k\rightarrow\infty}s(k) = (-1)^{k+1}(I+A)^{-1}b v_{n'+1}^T(1).
	\end{aligned}
\end{equation}
%
%
 In principle, from \eqref{eq:sk:infty}  
the value of $\lim_{k\to \infty} s(k)$ can be figured out once the value of $(I+A)^{-1}$ is determined.
However,
the direct calculation of inverse matrix is of high complexity.
Recall that the specific form of vector $b$ whose elements are all $0$ except for the last one. 
Thus the value of $(I+A)^{-1}b$ only depends on the last column of $(I+A)^{-1}$.
For the sake of simplified calculation, we focus on the recursive relationship in terms of the bottom right block of matrix $(I+A)^{-1}$.

In light of $A$ defined in \eqref{s:dynamics}, it follows 
\begin{equation*}
	I + A =  \left[\begin{array}{cc}
		2I & \Delta t * I\\
		\alpha A_{21} & I + A_{22}
	\end{array}\right].
\end{equation*}
Let $\beta \dfb \alpha \Delta t/2$. Then $(I+A)^{-1}$ can be written in the following block form 
\begin{small}
    \begin{equation}
        \begin{aligned}
            \left[\begin{array}{cc}
                2I & \Delta t * I\\
                \alpha A_{21} & I + A_{22}
            \end{array}\right]^{-1} = \left[\begin{array}{cc}
            * & * \\
            * & (I+A_{22}-\beta A_{21})^{-1}
            \end{array}\right]
        \end{aligned},
    \end{equation}
\end{small}
where $*$ represents some certain matrix of appropriate dimension. The invertibility of matrix $(I+A_{22}-\beta A_{21})$ is shown in Appendix \ref{invertibility}. Hence it follows from \eqref{eq:sk:infty} that $\lim_{k\to \infty} \|s(k)\|$ converges to a constant number. By recalling the fact that $\| v_{n'+1}(k+1) \|= \|v_{n'+1}(k) \|$, and $v_{n'}(k)$ comprises the stacked vector $s(k)$,  one knows $\lim_{k\to \infty}\| v_{n'}(k)\| /\|v_{n'+1}(k)\|$ is also a constant real number.

In the following contents, we use $M(d)$ to represent the leading principal submatrix of order $d$ of matrix $(I+A_{22}-\beta A_{21})$. 
Denote by $f(d)$ the last element in matrix $M^{-1}(d)$.
\begin{theorem}\label{thm3.6}
    Under controller \eqref{estimator}, the value of $n'$ can be inferred as followed:
    \begin{equation}\label{3.36}
        n'=\frac{\ln \bar{f}(n')-\ln \bar{f}(1)}{\ln \bar{f}(2)-\ln \bar{f}(1)}+1
    \end{equation}
    where
    $\bar{f}(d) = \frac{{f}(d) -\rho_1}{{f}(d) -\rho_2}$
    with $\rho_{1,2} = \frac{2(1+\beta)\pm 4\sqrt{\beta}}{(1-\beta)^2}$, and $f(1)$, $f(2)$ and $f(n')$ are given by
	$f(1) = \frac{1}{1+\beta}$, $f(2) = \frac{1+\beta }{(1+\beta)^2  - \frac{(1-\beta)^2}{4}}$ and  
    $f(n')=\lim_{k\rightarrow\infty}\frac{2\|v_{n'}(k)\| }{\|v_{n'+1}(k)\|}$.

\end{theorem}
\emph{Proof:}
	From the definition of matrix $A$ in \eqref{s:dynamics}, one gets the explicit form of matrix $(I+A_{22}-\beta A_{21})$ as 
\begin{equation*}
	\small
	\setlength{\arraycolsep}{1.5pt}
         \begin{bmatrix}
                1+\beta & \frac{1-\beta}{2} & 0 & \cdots& 0&0 \\
                \frac{1-\beta}{2} & 1+\beta & \frac{1-\beta}{2} & \cdots & 0& 0\\
                0 & \frac{1-\beta}{2} & 1+\beta & \cdots& 0& 0 \\
                \vdots & \vdots & \vdots & \ddots & \vdots & \vdots \\
                0&0&0&\cdots&1+\beta&\frac{1-\beta}{2}\\
                0&0&0&\cdots&\frac{1-\beta}{2}&1+\beta
        \end{bmatrix}.
\end{equation*}
Accordingly, the leading principal submatrices with $d\in \{1,2,n'\}$ are respectively in the form of 
$M(1) = 1 + \beta$, $M(2)= \left[\begin{array}{cc}
	1+\beta & (1-\beta)/{2}\\
	(1-\beta)/{2}&1+\beta
\end{array}\right]$, and $M(n') = (I+A_{22}-\beta A_{21})$. 
For any $1\leq d \leq n'$, there holds 
\begin{equation*}
	\small 	\setlength{\arraycolsep}{1.5pt}
	\begin{aligned}
		M(d) = \left[\begin{array}{ccc|c}
			& & &0\\
		  	& M(d-1) & &\vdots\\
			& & &(1-\beta)/2\\
			\hline
			0&\cdots&(1-\beta)/2&1+\beta.\\
		\end{array}\right]\\
	\end{aligned}
\end{equation*}
the inverse of which is 
\begin{equation*}
	\begin{aligned}
		M^{-1}(d) = \left[\begin{array}{cc}
			* & * \\
			* & (1+\beta  - \frac{(1-\beta)^2}{4}f(d-1))^{-1}
		\end{array}\right].
	\end{aligned}
\end{equation*}
Then $f(d)$ can be obtained in a recursive manner yielding 
\begin{equation}\label{3.30}
	\begin{aligned}
		f(d) = \frac{1}{1+\beta  - \frac{(1-\beta)^2}{4}f(d-1)}.
	\end{aligned}
\end{equation}
Two roots of the characteristic equation of \eqref{3.30} are $\rho_{1,2} = \frac{2(1+\beta)\pm 4\sqrt{\beta}}{(1-\beta)^2}$.
Recalling the definition of $\bar f(d)$, 
the general expression of the recurrence relation \eqref{3.30} is given by 
\begin{equation}\label{eq:fbar}
    \begin{aligned}
        \bar{f}(d) = \bar{f}(1)(\frac{\bar{f}(2)}{\bar{f}(1)})^{d-1}.
    \end{aligned}
\end{equation}
When $d=n'$, taking the natural logarithm on both sides of \eqref{eq:fbar} yields  
\begin{equation}\label{3.33a}
    \begin{aligned}
        n'=\frac{\ln \bar{f}(n')-\ln \bar{f}(1)}{\ln \bar{f}(2)-\ln \bar{f}(1)}+1.
    \end{aligned}
\end{equation}
In view of \eqref{eq:sk:infty}, the absolute value of $u_{n'}$ satisfies 
\begin{equation}\label{3.35a}
    \begin{aligned}
        \lim_{k\rightarrow\infty}\|v_{n'}(k)\| = \frac{1}{2}f(n')\|v_{n'+1}(k)\|.
    \end{aligned}
\end{equation}
Therefore it is straightforward to get 
\begin{equation}
    \begin{aligned}
        f(n')=\lim_{k\rightarrow\infty}\frac{2\|v_{n'}(k)\| }{\|v_{n'+1}(k)\|}.
    \end{aligned}
\end{equation}
This completes the proof. \hfill $\square$

\subsection{Estimation using the measurements from the last two time instant}
In this subsection, under the assumption that the measurements from the last two time instants are available, the controller for estimation is designed as
\begin{equation} \label{estimator2}
	\begin{aligned} 
		v_{0}(k+1)=&0 \\
		v_{i}(k+1)=&\frac{\alpha}{2}\left(q_{i+1}(k)+q_{i-1}(k)-2 q_{i}(k)\right)\\&+\frac{v_{i+1}(k-1)+v_{i-1}(k-1)}{2}, i \in \mathcal{V}_C-\{0\} \\
		v_{n}(k+1)=&-v_{n}(k). \\
	\end{aligned}
\end{equation}
This controller is similar to \eqref{estimator} except that for $i\in \mathcal{V}_C-\{0\}$ it uses  $v_i(k-1)$ instead of $v_i(k)$.   
This specific manner contributes to {\color{black}an} analysis-friendly structure that will be illustrated below. 
Denote by $s_{r}(k) = [q_1(k), q_2(k),\ldots,q_{n'}(k), v_1(k-1), v_2(k-1),\ldots, v_{n'}(k-1), v_1(k), v_2(k), \ldots, v_{n'}(k)]^T$.
The compact form of (\ref{estimator2}) is
\begin{equation*} \label{3.35}
	\begin{aligned}
		\small
		s_r(k+1) =
		\underbrace{\begin{bmatrix}
			I & \mathbf{0} & \Delta t * I \\
			\mathbf{0} & \mathbf{0} &I\\
			\alpha A_{21} & A_{22}& \mathbf{0}\\  
        \end{bmatrix}}_{\dfb A_r} s_r(k) + bv^T_{n'+1}(k) \\
	\end{aligned}
\end{equation*}
The definition of $A_{21}$, $A_{22}$ and $b$ are the same as that in Subsection \ref{sec:3a}.
Similarly, one has 
\begin{equation*}
	\begin{aligned}
		s_r(k+1)=&A_r^ks(1) + \sum_{i=1}^k\left(A_r^{k-i}bv_{n'+1}^T(i)\right)\\
		= &A_r^ks(1)  +(-1)^{k+1}\sum_{i=1}^k\left((-A_r)^{k-i}bv_{n'+1}^T(1)\right).
	\end{aligned}
\end{equation*}
Then we have another main theorem regarding the spectral property of matrix $A_r$. 
\begin{theorem}\label{lem2.1}
    {\color{black}The spectral radius of $A_r$ is less than $1$ if the parameter is chosen such that $\alpha \Delta t < \frac{1-\cos^2(\frac{\pi}{n'+1})}{5+\cos^2(\frac{\pi}{n'+1})}$.}
\end{theorem}
The proof of Theorem \ref{lem2.1} is given in Appendix \ref{app2}.

Following the same operations as the previous subsection, one has 
\begin{equation}\label{3.43}	
		\lim_{k\rightarrow\infty}s_r(k) = (-1)^{k+1}(I+A_r)^{-1}b v_{n'+1}^T(1).	
\end{equation}
\begin{theorem}\label{them:4}
    Under controller \eqref{estimator2}, the value of $n'$ can be obtained in the form of 
    \begin{equation}\label{3.55}
		\begin{aligned}
			n'=\lim_{k\rightarrow\infty}\frac{(1-\beta)\|v_{n'}(k)\| }{\|v_{n'+1}(k)\| - (1-\beta) \|v_{n'}(k)\| }.
		\end{aligned}
	\end{equation}
\end{theorem}
\emph{Proof:}
The inverse of matrix $(I+A_r)$ is given by 
\begin{equation*}
	\begin{aligned}
		(I+A_r)^{-1}
&=\left[\begin{array}{cc}
	* & * \\
	* & (I-A_{22}-\beta A_{21})^{-1}
\end{array}\right]
	\end{aligned}
\end{equation*}
where the explicit form of $I-A_{22}-\beta A_{21}$ is 
\begin{equation*}
	\small
	\setlength{\arraycolsep}{1.5pt}
        \begin{bmatrix}
                1-\beta & (1-\beta)/{2} & \cdots& 0&0 \\
                (1-\beta)/{2} & 1-\beta   & \cdots & 0& 0\\
                0 &  (1-\beta)/{2} &  \cdots& 0& 0 \\
                \vdots & \vdots &  \ddots & \vdots & \vdots \\
                0&0&\cdots&1-\beta& (1-\beta)/{2} \\
                0&0&\cdots& (1-\beta)/{2} &1-\beta
        \end{bmatrix}.
\end{equation*}

To distinguish from the symbol $M(d)$ in previous subsection, we use $M_r(d)$ to denote the leading principal submatrix of order $d$ of matrix $(I-A_{22}-\beta A_{21})$.
Then its determinant can be obtained via
\begin{equation*}
	|M_r(d)| = (1-\beta)|M_r(d-1)| - \frac{(1-\beta)^2}{4}|M_r(d-2)|. 
\end{equation*}
The general expression of $M_r(d)$ from the above recursive equation is given by 
\begin{equation*}
	\begin{aligned}
		|M_r(d)| = \frac{n+1}{2^n}(1-\beta)^n
	\end{aligned}.
\end{equation*}
Apparently, when $\beta \neq1$, $|M_r(d)|\neq 0,\forall d\in \mathbb{N}$, implying $M_r(d)$ is invertible. 
Let $g(d)$ represent the last element of matrix $M_r^{-1}(d)$.
Then it follows
\begin{equation*}
	\begin{aligned}
		M_r^{-1}(d) = \left[\begin{array}{cc}
			* & * \\
			* & \left(1-\beta  - \frac{(1-\beta)^2}{4}g(d-1)\right)^{-1}
		\end{array}\right].
	\end{aligned}
\end{equation*}
The function $g(d)$ can be expressed as a recurrence relation 
\begin{equation*}
	\begin{aligned}
		g(d) = \frac{1}{1-\beta  - \frac{(1-\beta)^2}{4}g(d-1)},
	\end{aligned}
\end{equation*}
the explicit solution to which is given by 
\begin{equation*}
		g(d) = \frac{2d}{(d+1)(1-\beta)}.
\end{equation*}
In combination with \eqref{3.43}, as $k\to \infty$, $v_{n'}(k)$ satisfies 
\begin{equation}
	\begin{aligned}
		\lim_{k\rightarrow\infty}\|v_{n'}(k)\| = \frac{n'}{(n'+1)(1-\beta)}\|v_{n'+1}(k)\|.
	\end{aligned}
\end{equation}
Then after simple rearranging, the value of $n'$ can be obtained as \eqref{3.55}.  \hfill $\square$

\begin{remark}\label{rem:01}
	Note that although the two calculation manner \eqref{3.36} and (\ref{3.55}) both require the iterative step $k$ tends to infinity, in implementations and applications the value of $n'$ can be obtained in finite time.
	Since the eventual estimation value of $n'$ is a positive integer, $n'(k)$ will not be updated once $n'(k)$ enters the interval of $(-0.5,0.5)$ around some constant value. The real value can then be obtained via rounding-off method.
\end{remark}

\section{FORMATION CONTROL BASED ON ESTIMATION}\label{formation}
This section will present control law for each robot based on the estimation of robot number in each chain. 
Given that the vertex robot $s_i$  has the knowledge of $n^s_{i-1}$ via estimation, the polygon formation control law is designed as 
\begin{small}
    \begin{equation}\label{eq:control}
        \begin{aligned}
            v_{i}(k+1)=&\frac{\alpha}{2}\left(q_{i+1}(k)+q_{i-1}(k)-2 q_{i}(k)\right)\\
            &+\frac{v_{i+1}(k+1-\sigma_k)+v_{i-1}(k+1-\sigma_k)}{2}, i \in \mathcal{V}-S\\
            v_{i}(k+1)=&\alpha\left(q_{i-1}(k)-q_{i}(k) - l_{i-1}^* \right)\\
            &+v_{i-1}(k+1-\sigma_k) ,i\in{S}-\{s_0\}\\
            v_{s_{0}}(k+1) =& 0
        \end{aligned}
    \end{equation}
\end{small}
where $\sigma_k\in\{1,2\}$ indicates the time instants associated with the measurements used in implementation and denote $l_{i-1}^*=\frac{r^*_{i-1}}{n^s_{i-1}}$. 
{\color{black}It can be observed from \eqref{eq:control} that the external information $l^*_{i-1}$ only influence the vertex robots, while for the non-vertex robot, the controllers of the estimation and the formation process share the same form.
Hence, the two processes can be implemented successively.}

\begin{theorem}\label{thm:control}
	   Using the control law \eqref{eq:control}, the group robots modelled by \eqref{sys:model} are stabilized at the desired polygon formation under the parameter condition in Theorem \ref{lem3.1}.      
\end{theorem}

\emph{Proof:}
The proof is divided into three steps: a) clarify the compact form of the system under control law \eqref{eq:control}; 
b) prove the Schur stability of the state matrix; 
c) show the convergence to the desired state.

Firstly, according to \eqref{eq:control}, the entire system is a linear cascade system where every two chains $s_i,s_{i-1}\in S$ are cascaded with $q_{s_{i-1}}$ and $v_{s_{i-1}}$.
For the sake of brevity, suppose the number of robots in each chain are all equal to $n$.
Similarly, the dynamics under the formation controller \eqref{eq:control} when $\sigma_k = 1$ can be written as 
\begin{equation}\label{s:formation_dynamics}
	\begin{aligned}
		s(k+1)=\underbrace{\begin{bmatrix}
			I & \Delta t * I\\
			\alpha A_{21f} & A_{22f}   
        \end{bmatrix}}_{\triangleq A_f} s(k)
		+ B_fu_f,
	\end{aligned}
\end{equation}
where $A_{21f},A_{22f}\in\mathbb{R}^{n\times n}$,
\begin{equation*}
	\small
	B_f = 
	\begin{bmatrix}
		\mathbf{0}_{n\times 3} \\
		\begin{matrix}
			0.5\alpha & 0.5 & 0 \\
			0 & 0 & 0 \\
			\vdots & \vdots & \vdots \\
			0 & 0 & -\alpha
		\end{matrix}
	\end{bmatrix}
\end{equation*}
and $u_f = [q_{s_{i-1}}, v_{s_{i-1}}, l_{i-1}^*]^T$.
By applying iterative process, \eqref{s:formation_dynamics} turns to be 
\begin{equation*}
	s(k+1) = A_f^ks(1)+(I+A_f+...+A_f^{k-1})B_fu_f.
\end{equation*}

Secondly, we prove that the state matrix is Schur, i.e., $\lim_{k\rightarrow \infty} A_f^k= 0$.
Noticing that the matrix $A$ and $A_f$ only differ in two entries,
we separate $A_f$ into $A_f = A + A_d$ with $$A_d =
 \begin{bmatrix}
	\mathbf{0}_{(2n-1)\times n} & \mathbf{0}_{(2n-1)\times n} \\
	\begin{matrix}
		0 & \cdots & 0.5\alpha & 0
	\end{matrix} &
	\begin{matrix}
		0 & \cdots & 0.5 & 0
	\end{matrix}
\end{bmatrix}.$$
Then, 
\begin{equation}\label{sep}
	A_f^k = (A+A_d)^k = A^k + kA^{k-1}A_d,
\end{equation} 
since $A_d^k=0,k=2,3,...,\infty$.
Remind that we already have $\lim_{k\rightarrow \infty} A^k= 0$ when the parameters satisfy Theorem 1.
Now we focus on the second term.
The Jordan normal form of $A$ can be obtained as $A = PJP^{-1}$.
Assume that $\lambda$ is an eigenvalue of $A$ and $J(\lambda)\in \mathbb{R} ^{n_\lambda\times n_\lambda}$ is its corresponding Jordan block with $n_\lambda$ being the dimension of $J(\lambda)$.
We have
\begin{equation*}
	\small 
	J(\lambda)^{k-1}=
	\begin{bmatrix}
		\lambda ^{k-1} & (k-1)\lambda^{k-2} & ... & C^{k-2}_{n_\lambda}\lambda^{k-n_\lambda} \\
			0     &  \lambda^{k-1}       & ... & C^{k-3}_{n_\lambda}\lambda^{k-n_\lambda+1} \\
			 \vdots & \vdots & \ddots & \vdots \\
			0 & 0 & ... & \lambda^{k-1}
	\end{bmatrix}.
\end{equation*}
As $|\lambda|<1$, it is obvious that $\lim_{k\rightarrow \infty }kJ(\lambda)^{k-1}=0$,
which implies that $\lim_{k\rightarrow \infty }kA^{k-1}A_d = \lim_{k\rightarrow \infty }kPJ^{k-1}P^{-1}A_d = 0$.
Combining \eqref{sep}, we have $\lim_{k\rightarrow \infty }A_f^k=0$ when $\alpha \Delta t < \frac{1-\cos^2(\frac{\pi}{n+1})}{3-\cos^2(\frac{\pi}{n+1})}$.
Further, the matrix of the whole system \eqref{eq:control} is a lower triangular matrix, denoted by 
\begin{equation*}
	\small 	
	A_s=
	\begin{bmatrix}
		A_f  &   \mathbf{0}_{2n\times 2n}  & \mathbf{0}_{2n\times 2n}&  \cdots   & \mathbf{0}_{2n\times 2n} \\
        A_{sd} &
		A_f   & \mathbf{0}_{2n\times 2n} &  \cdots  & \mathbf{0}_{2n\times 2n}\\	
		\mathbf{0}_{2n\times 2n} & A_{sd}  & A_f & \cdots & \mathbf{0}_{2n\times 2n}\\
		\vdots & & \ddots & & \vdots \\
		\mathbf{0}_{2n\times 2n} & \mathbf{0}_{2n\times 2n} & \cdots & A_{sd}  & A_f \\
	\end{bmatrix}
\end{equation*}
with
\begin{equation*}
	A_{sd} = \begin{bmatrix}
		\mathbf{0}_{n\times n} &  \mathbf{0}_{n\times n}   \\
		\begin{matrix}
			0 & 0 & \cdots & 0.5\alpha 
		\end{matrix} &
		\begin{matrix}
			0 & 0 & \cdots & 0.5\alpha 
		\end{matrix} \\
		\mathbf{0}_{(n-1)\times n} &  \mathbf{0}_{(n-1)\times n} 
	\end{bmatrix}.
\end{equation*}
Therefore, the whole system matrix has the same eigenvalue as $A_f$, which implies the whole system is stabilized.
If the number of robots in each chain is different, the stabilization condition is up to the largest $n_{i-1}^s$.

Finally, we prove that the system converges to the desired state under the control input $u_f$.
In view of the fact that $\lim_{k\rightarrow \infty }A_f^k=0$ implies the spectral radius of $A_f$ is less than 1\cite{matrixanalysis},
it yields
\begin{equation}\label{formation_con}
	\lim_{k\rightarrow \infty }s(k)=(I-A_f)^{-1}B_fu_f,
\end{equation}
where
\begin{equation*}
	(I-A_f)^{-1} = 
	\begin{bmatrix}
		(\alpha \Delta t)^{-1}I & -\alpha^{-1}A_{21f}^{-1} \\ 
		-\Delta t^{-1}I  &  \mathbf{0}
	\end{bmatrix}.
\end{equation*}
The value of $(I-A_f)^{-1}B_fu_f$ only depends on the first and the last column of $A_f$.
Notice that the first column of $A_{21f}^{-1}$ is $[-2,-2,...,-2]^T$ and the last column is $[-1,-2,...,-n]^T$.
Then \eqref{formation_con} turns to 
\begin{equation*}
	\begin{aligned}
		\lim_{k\rightarrow \infty }&s(k) 
		= [q_{s_{i-1}}+\alpha^{-1}v_{s_{i-1}}-l_{i-1}^*,q_{s_{i-1}}+\alpha^{-1}v_{s_{i-1}}\\
		&-2l_{i-1}^*,..., 
		 q_{s_{i-1}}+\alpha^{-1}v_{s_{i-1}}-nl_{i-1}^*, 0,0,...,0]^T.
	\end{aligned}
\end{equation*}
For the first chain, it is set that $q_{s_0}=q_0$ and $v_{s_0} = 0$ where $q_0$ is an arbitrary desired position. 
Then, the convergent position of the first chain is $[q_0-l_0^*, q_0-2l_0^*,...,q_0-nl_0^*]$ and the ultimate velocity is $0$.
Similarly, the convergence position of the second chain is $[q_0-r_0^*-l_1^*, q_0-r_0^*-2l_1^*,...,q_0-r_0^*-nl_1^*]$. 
The convergence state of the succeeding chains can be deduced in the same way, indicating the whole system will converge to the desired state.
The proof of the case when $\sigma_k = 2$ is quite similar and is omitted due to the space limitation.

\section{SIMULATIONS AND EXPERIMENTS}\label{simulation}
In this section, we first present the simulation results to validate the effectiveness of the two estimation strategies. Their performance in terms of the convergence speed and the sensitivity to robot group size will also be discussed. Then the simulation and experimental results are presented to give an intuitive sense on the behavior of the proposed control scheme.

\subsection{Simulation results of estimation strategies}
The simulation is conducted with $20$ robots that are randomly distributed on a chain graph. 
The time interval between two sampling instants is set to be $\Delta t=0.01s$ and the parameter $\alpha$ in different controllers are chosen to be the same as $\alpha =0.5$.
\begin{figure}
	\centering
	\includegraphics [width=0.3\textwidth]{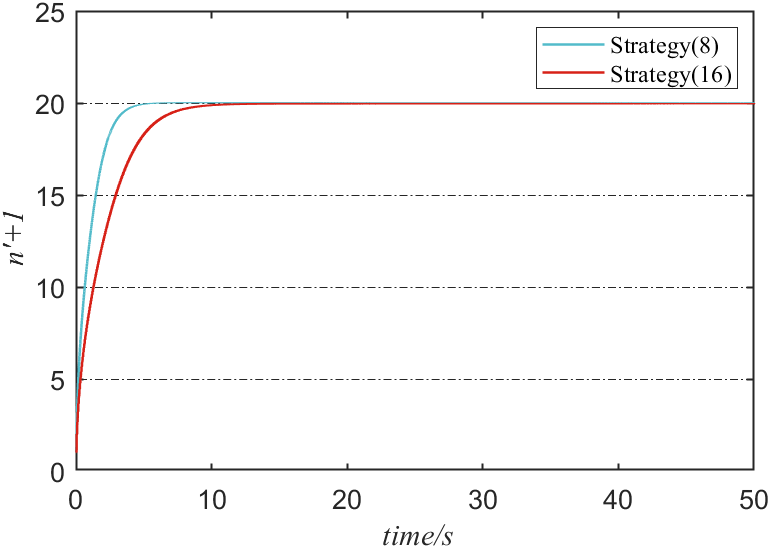}
	\caption{Estimation value over time using \eqref{3.36} and \eqref{3.55}}\label{fig:3disEst}
\end{figure}
\begin{figure}
	\centering
	\includegraphics [width=0.3\textwidth]{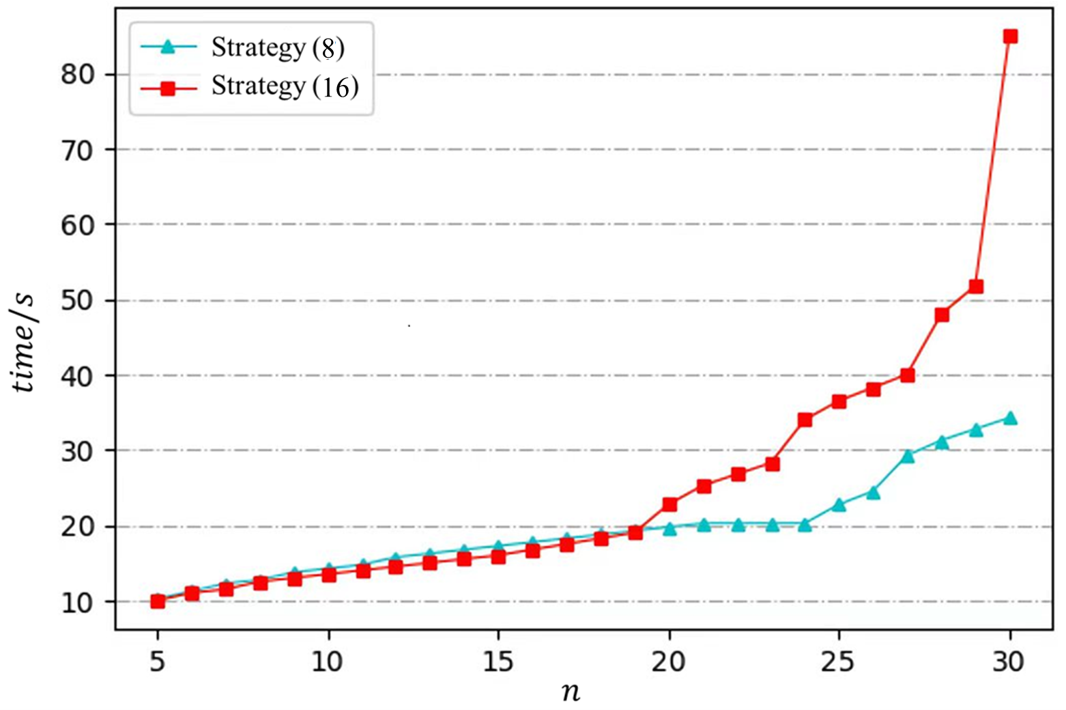}
	\caption{Convergence time under different numbers of robots }\label{fig:3covtime}
\end{figure}
\begin{figure}
	\centering
	\includegraphics [width=0.3\textwidth]{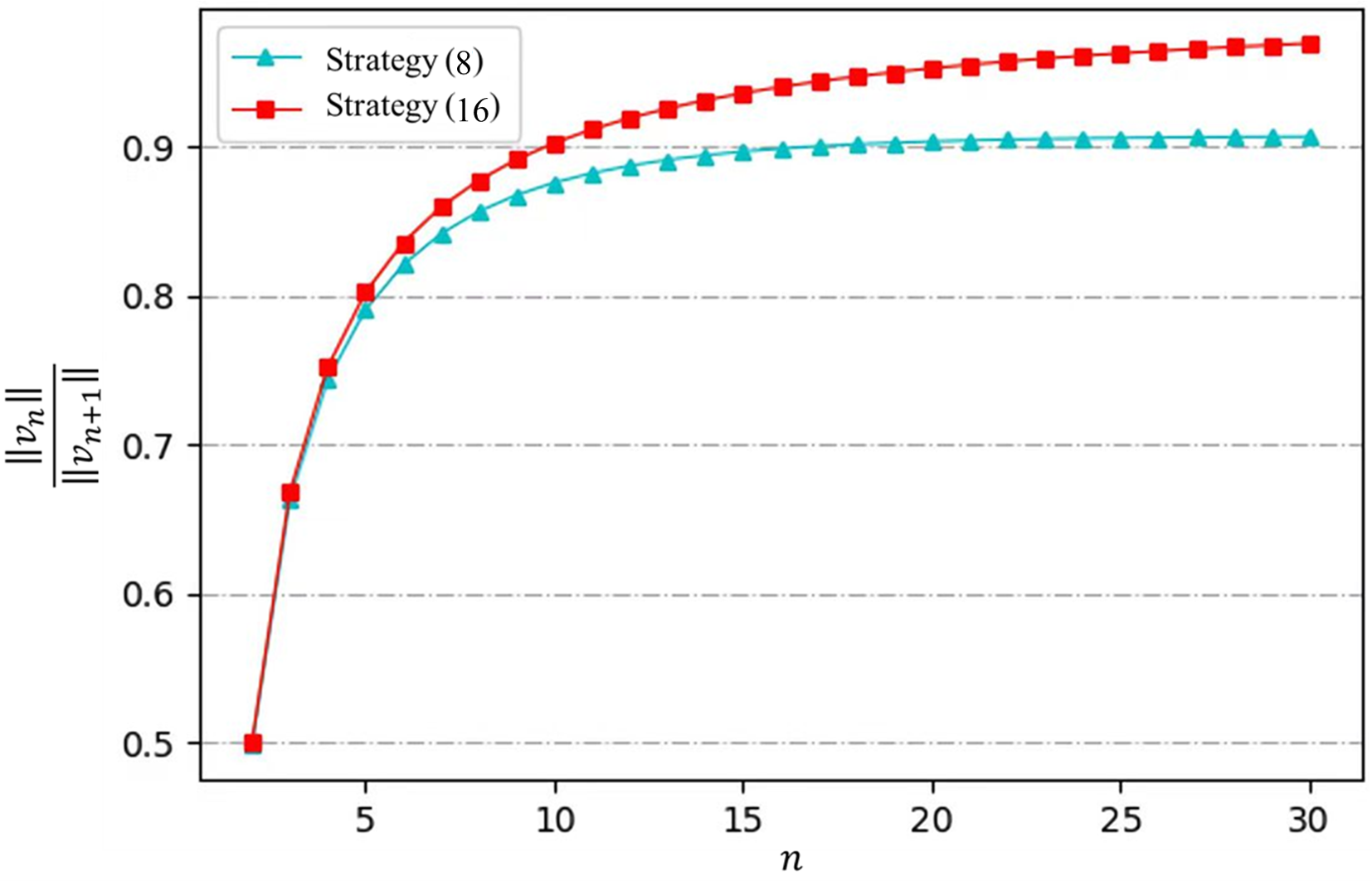}
	\caption{Sensitivity test to the number of robots}\label{fig:3covval}
\end{figure}

Fig. \ref{fig:3disEst} shows the evolution of estimation value using estimation strategies \eqref{3.36} and \eqref{3.55} over time, from which it can be seen the precise estimation can be achieved in finite time. 
{\color{black}Besides the effectiveness, we compare the two estimation strategies from the perspective of their convergence speed, the robustness, and the computation complexity. The comparison of convergence speed is carried out by setting the number of robots from $5$ to $30$, }and recording the convergence time at each $n$. We then derive the average time after repeating the same operation five times. 
The results are shown in Fig. \ref{fig:3covtime}.
It can be observed that when the group size is relatively small, 
the convergence speed is almost the same no matter which strategy is used. However as the size of the robot group grows, the strategy \eqref{3.36} renders us precise estimation in less time than \eqref{3.55}. In addition, from the explicit expressions of \eqref{3.36} and \eqref{3.55}, we know the precise estimation relies on both $\|v_{n'}\|$ and $\|v_{n'+1}\|$ when they reach their equilibrium. In order to show the influence of group size on estimation, we conduct another simulation by computing the change of $\|v_{n'}\|/\|v_{n'+1}\|$, which can be interpreted as the sensitivity (or somewhat robustness) w.r.t. the number of robots. 
{\color{black}The results are shown in Fig. \ref{fig:3covval}, implying the strategy (16) is more sensitive to the group size, which is more favorable to the estimation. }
It is also worth noticing that irrespective of those above-mentioned properties, the relatively more concise expression of \eqref{3.55} generally leads to lower computation complexity.

\subsection{Simulation of formation control}

{\color{black}
Consider a team of $120$ robots whose desired formation is a hexagon, with $20$ robots on each chain.
The set of vertex robots is set as $S=\{1,21,41,61,81,101\}$ and the corresponding relative configuration $r^*$ is chosen as 
\begin{equation*}
	\begin{aligned}
	r^*=\left[\begin{array}{cccccccc}{-4} & {-8} & {-4} & {4} & {8} & {4} \\ {-8} & {0} & {8} & {8} & {0} & {-8}\end{array}\right].
	\end{aligned}
\end{equation*}
Assume that the formation control law \eqref{eq:control} is implemented under the condition that robot $s_i$ has obtained the real value of $n^s_{i-1}$ via estimation. 
The time interval is set to be $\Delta t=0.05s$ and the control parameter $\alpha =0.5$.
\begin{figure}
	\centering
	\subfigure [$t=0s$]{
		\includegraphics [width=0.18\textwidth]{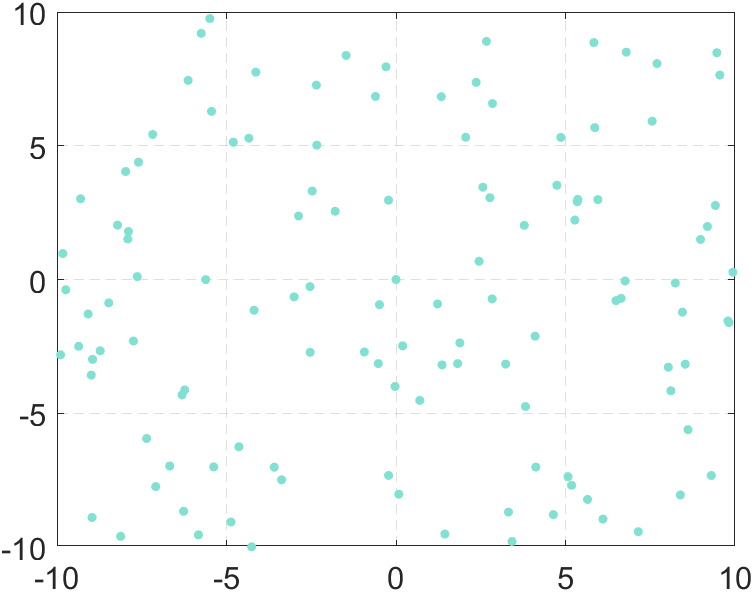}}
	\subfigure [$t=50s$]{
		\includegraphics [width=0.18\textwidth]{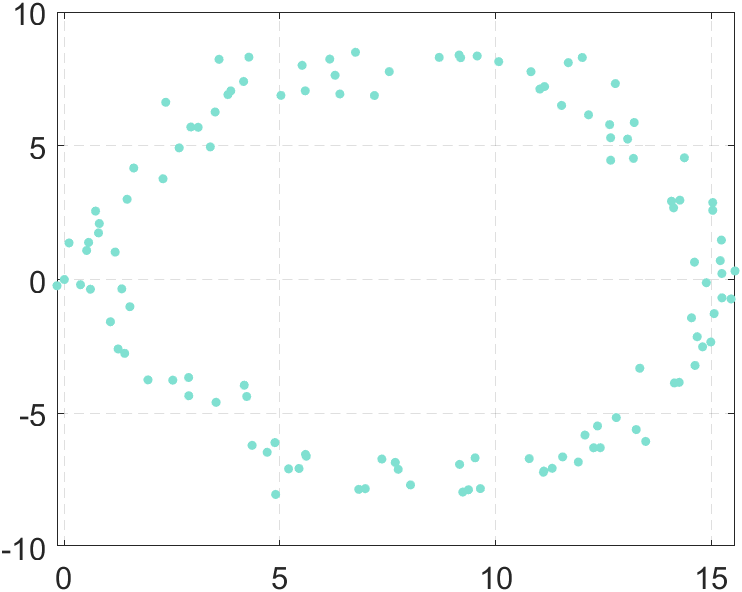}}
	\subfigure [$t=100s$]{
		\includegraphics [width=0.18\textwidth]{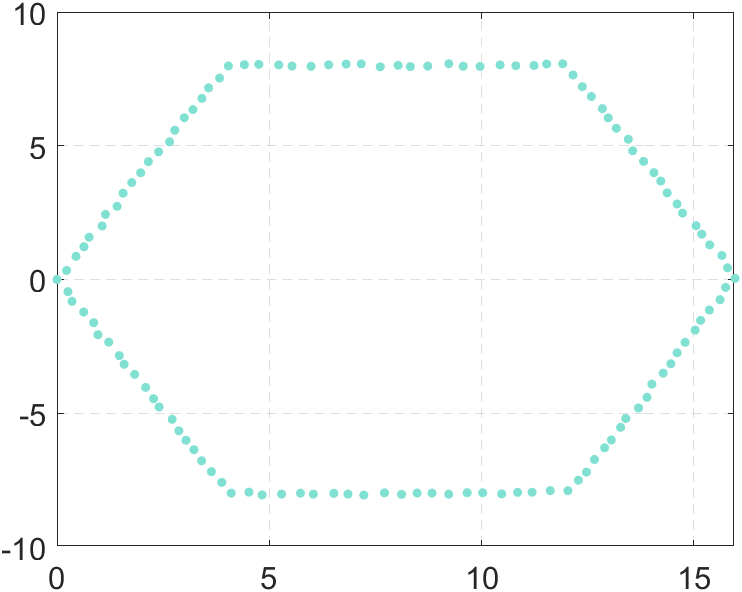}}
	\subfigure [$t=150s$]{
		\includegraphics [width=0.18\textwidth]{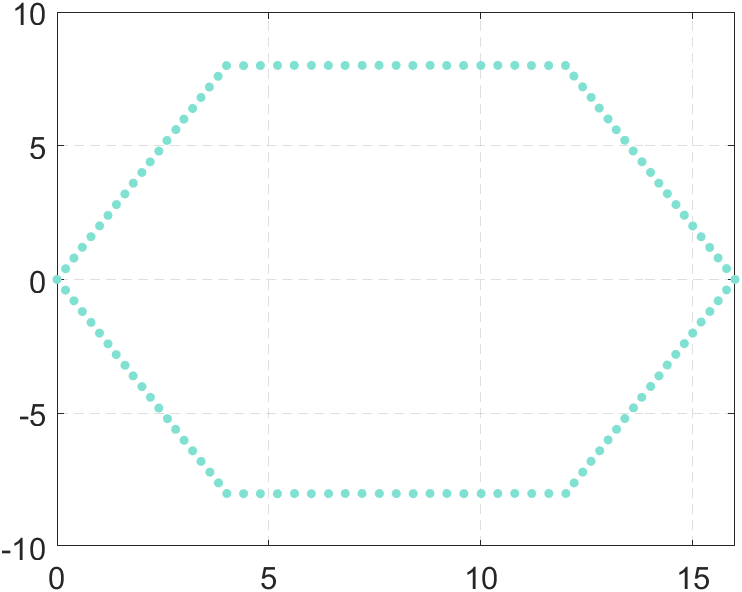}}
	\caption{The snapshots of $120$ robots converging to the desired formation under the controller \eqref{eq:control}}\label{fig:3forpos}
	\label{fig:3form}
\end{figure}
\begin{figure}
	\centering
	\includegraphics [width=0.28\textwidth]{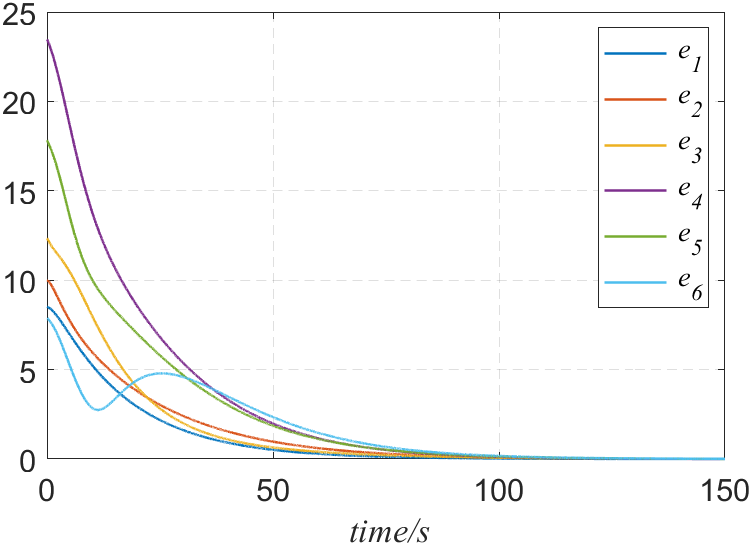}
	\caption{The relative distance errors w.r.t. neighboring vertex robots }\label{fig:3poserr}
\end{figure}
\begin{figure}
	\centering
	\includegraphics [width=0.28\textwidth]{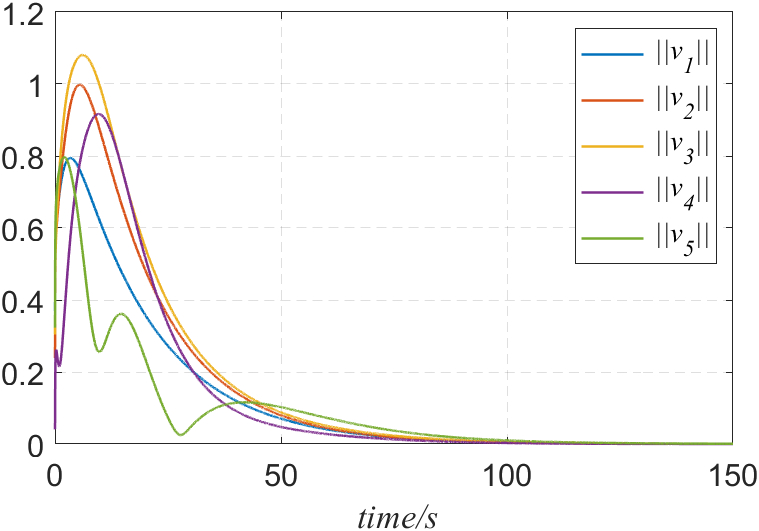}
	\caption{\color{black}The velocity of each vertex robot}\label{fig:3forvel}
\end{figure}
Fig. \ref{fig:3forpos} shows the collective formation shape at $t=\{0, 50, 100, 150\}$s. Based on the formation evolution at different time instants, it is obvious that the the desired formation is achieved from the geometric perspective.  
This is further validated by the convergence of relative distance errors $e_i \dfb \|r_{i} - r_i^*\|, i=1,\cdots,5$, to the origin, shown in Fig. \ref{fig:3poserr}. When equilibrium is attained, the robots become static and maintain the status thenceforth, which is demonstrated in Fig. \ref{fig:3forvel}.  
}

\subsection{Experiments}

\begin{figure}
	\centering
	\includegraphics [width=0.3\textwidth]{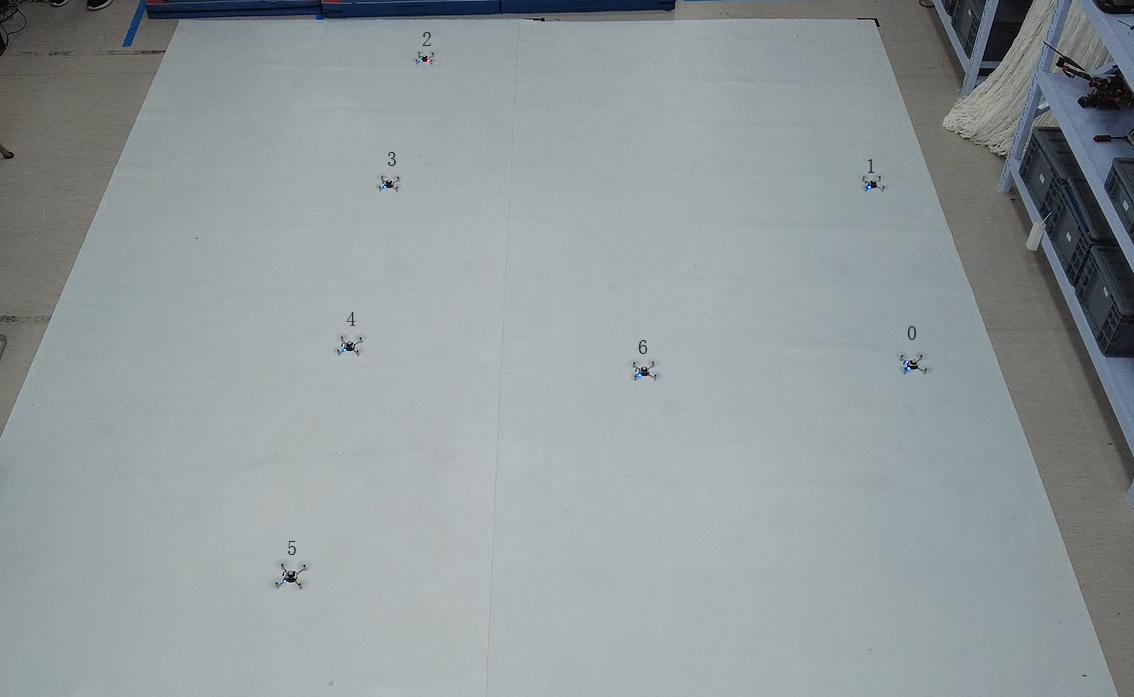}
     \caption{Initial configuration of the robot team}
    \label{estimatesnap}
\end{figure}

\begin{figure}[htbp]
    \centering
    \subfigure[algorithm \eqref{3.36}]{
    \begin{minipage}[t]{0.45\linewidth}
    \centering
    \includegraphics[width=1.6in]{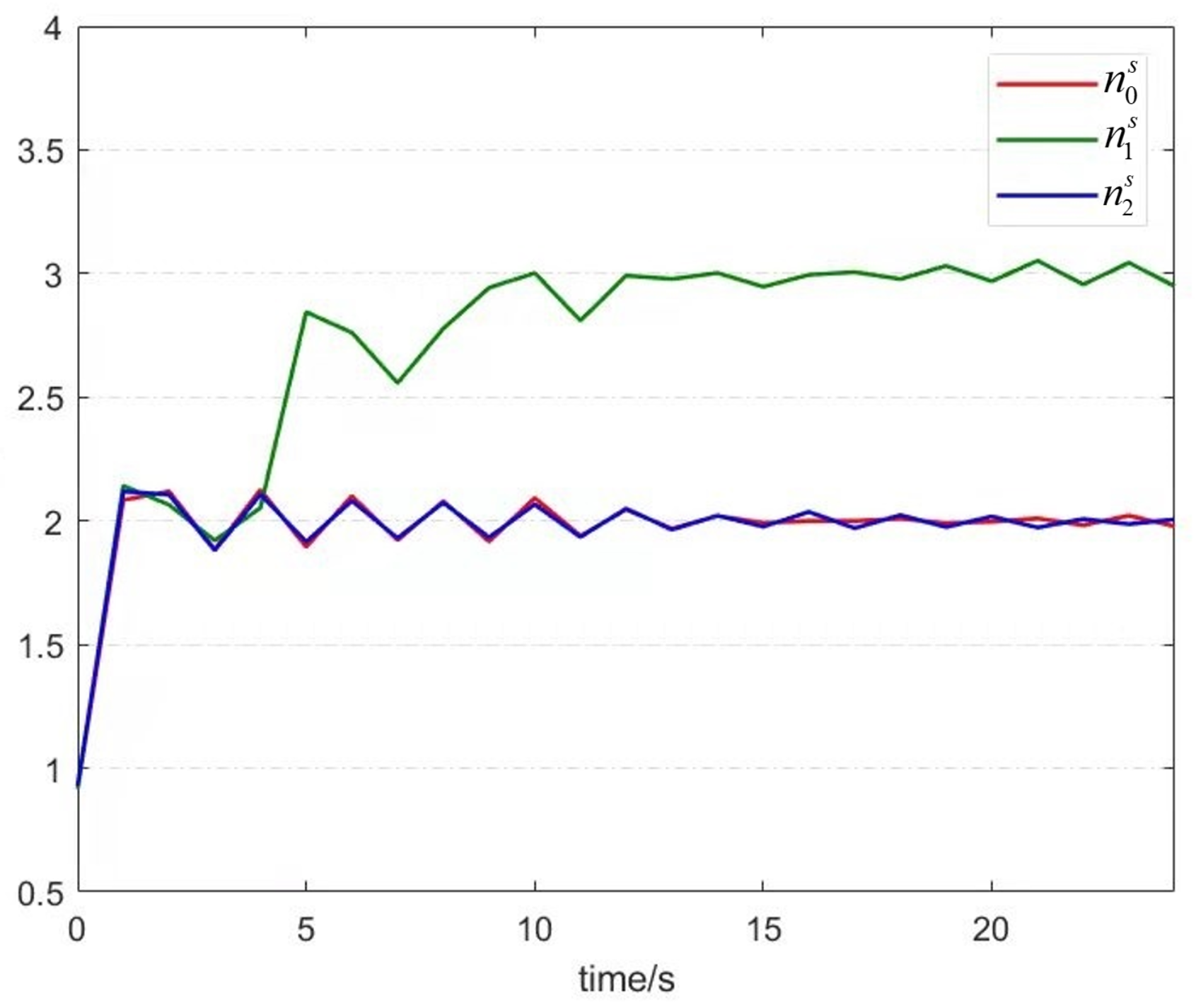}
    \end{minipage}%
    }%
    \subfigure[algorithm \eqref{3.55}]{
    \begin{minipage}[t]{0.45\linewidth}
    \centering
    \includegraphics[width=1.6in]{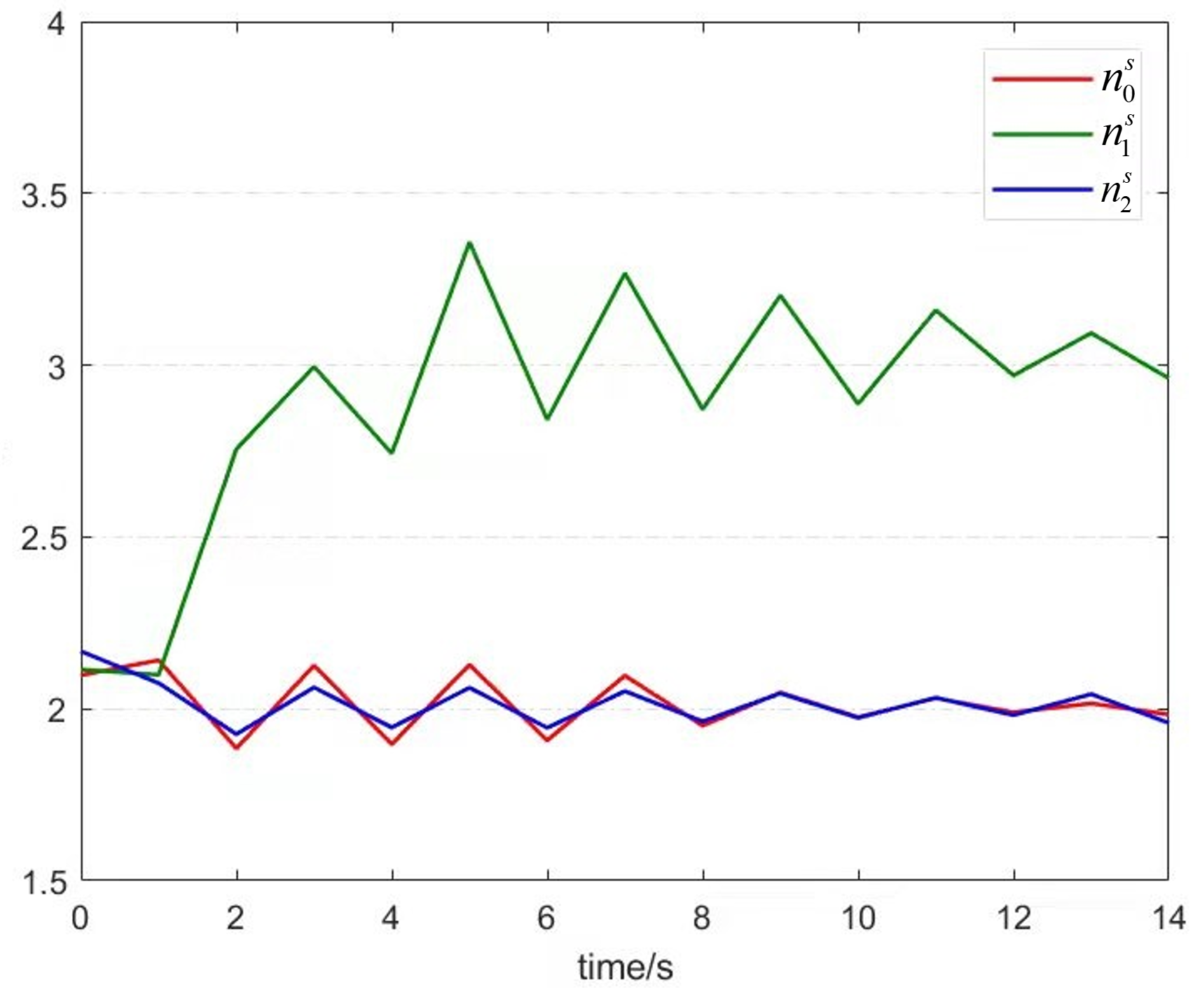}
    \end{minipage}%
    }%
    \centering
    \caption{Estimation of the number of robots in three chains using two different algorithms}
    \label{experiment}
\end{figure}

In this subsection, the physical experiments are carried out on the mobile platform consisting of 7 miniature unmanned aerial vehicles called Crazyflie. 
{\color{blue}The Crazyflie is a typical quadrotor UAV. Generally, the controller is designed in cascade form with two sub-controllers: an inner-loop attitude controller and an outer-loop position controller. We only focus on the latter, where the kinematics can be described by \eqref{sys:model}.}
Two phases are involved: distributed estimation and formation control. 
The initial relative locations of these flying robots are shown in Fig. \ref{estimatesnap}. The desired polygon formation is prescribed as a triangle with vertex robot set $S=\{0,2,5\}$. Hence three chain graphs are accordingly generated, containing $2$, $3$ and $2$ robots respectively. 
In this situation, the time interval is set to be $\Delta t=1s$ and the control parameter $\alpha =0.1$.
Fig. \ref{experiment} shows the results of estimations implementing algorithms \eqref{3.36} and \eqref{3.55}. It is easy to see the precise estimation can be obtained using either of them. 

\begin{figure}[htbp]
    \centering
    \subfigure[The stabilized formation]{
    \begin{minipage}[t]{0.5\linewidth}
    \centering
    \includegraphics[width=1.65in]{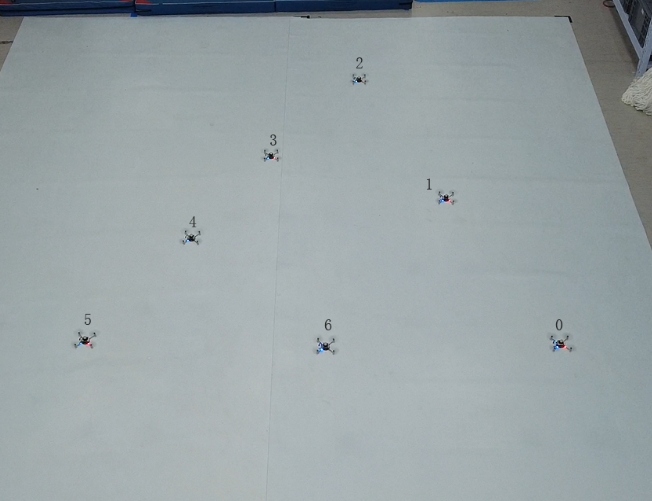}
    \end{minipage}%
    }%
    \subfigure[The relative distance errors]{
    \begin{minipage}[t]{0.5\linewidth}
    \centering
    \includegraphics[width=1.6in]{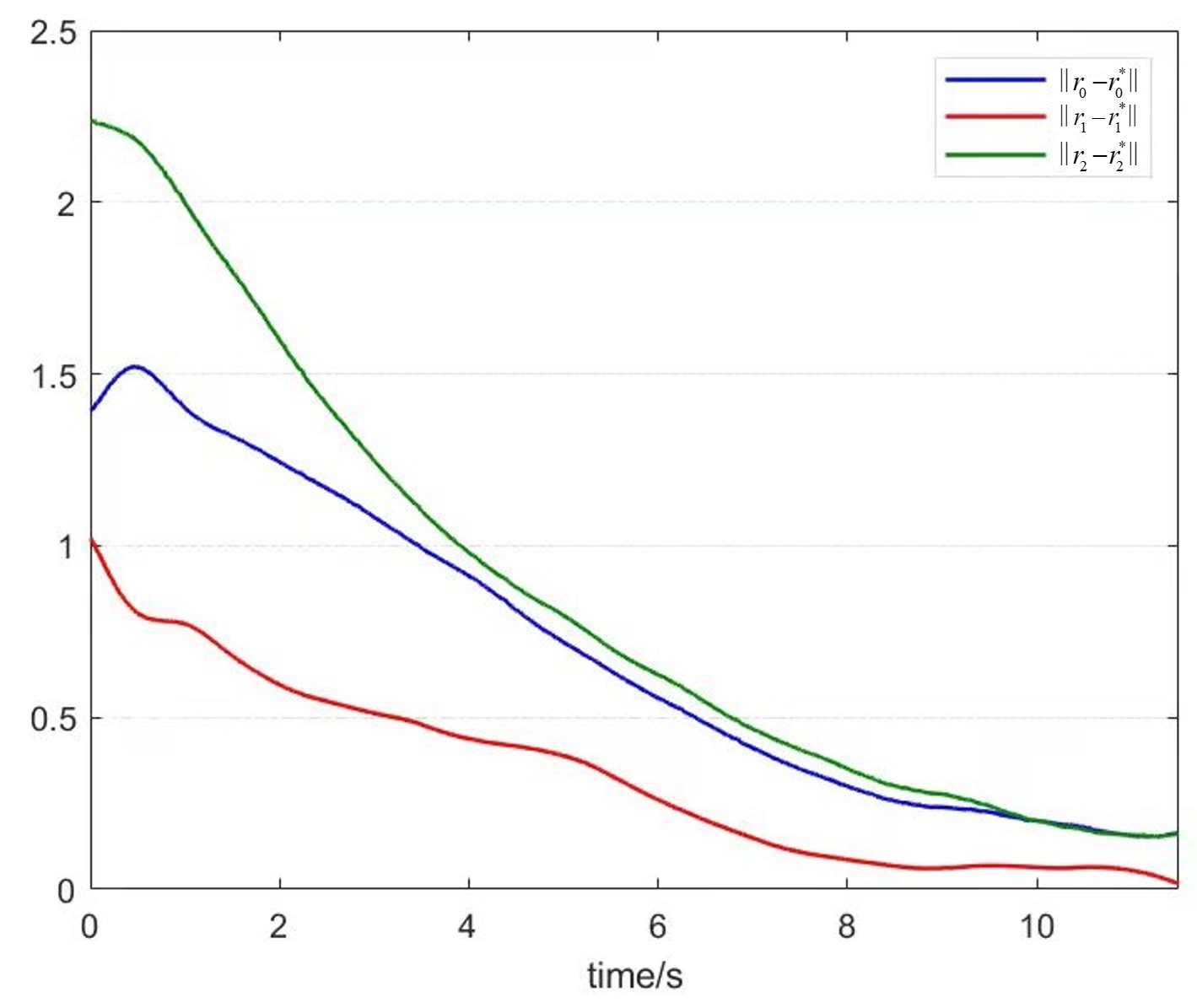}
    \end{minipage}%
    }%
    \centering
    \caption{Final configuration of the robot team}
    \label{final-configuration}
\end{figure}

    In formation control, the relative position matrix of neighboring vertex robots is designed as 
\begin{equation*}
    r^*=
    \begin{bmatrix}
        1 &  2 &  -3  \\
        -2 & 2 & 0
    \end{bmatrix}.
\end{equation*}
The parameters are chosen as $\Delta t=0.2s$ and $\alpha =0.3$. After implementing the control law \eqref{eq:control}, the robots are stabilized at a triangle formation shown in Fig. \ref{final-configuration}(a), where the in-between robots are evenly distributed along each side. Regarding the vertex robots, their relative distance errors are shown in Fig. \ref{final-configuration}(b), where the the convergence to the origin indicates the realization of the prescribed polygon formation shape. Together with the previous discussion on the rest robots, the effectiveness of the proposed self-organized formation control strategy is verified via physical flying robots.

\section{Conclusion}
\label{conclusion}
In this paper,
we have proposed a self-organized polygon formation control framework
that can realize an arbitrary polygon formation with given vertex robots.
Firstly, two distributed control strategies for estimation have been designed using 
the measurements from the latest and the last two time instants respectively.
Based on the estimation, the vertex robots can infer the number of robots in its associated chain.
Then under the circumstance that only vertex robots have access to the external information, the specific formation control law has been proposed for each robot so as to enable the majority of the group robots move merely following the very simple principle, namely moving towards the centroid of the line segment formed by two direct neighbors.  
The proposed polygon formation strategy extricates the users  
from complicated pre-design of the desired relative variables globally.
In addition, it is inherently superior to the consensus-based control structure due to its scalability and flexibility in the sense that the external information only relates to a few robots. 
An interesting direction in the future is to extend the polygon formation to more general formation shapes. 

\section*{Appendix}
\subsection{Proof of Theorem 1}\label{app1}
    The characteristic polynomial of $A$ is
    \begin{equation*}
        \begin{aligned}
            |\lambda I-A|&=\left|\begin{array}{cc}
                (\lambda-1)I & -\Delta t * I \\
                -\alpha A_{21} & \lambda I-A_{22}
            \end{array}\right|\\
            &=\left|(\lambda-1)(\lambda I-A_{22})-\Delta t\alpha A _{21}\right|,\\
        \end{aligned}
    \end{equation*}
    where $\lambda\in\mathbb{C}$.
    Let $\bar{A}\dfb (\lambda-1)(\lambda I-A_{22})-\Delta t\alpha A_{21}$, the explicit form of which is 
        \begin{equation*}
            \bar{A}=
            \begin{bmatrix}
                \begin{smallmatrix}
                    \lambda^2-\lambda+\alpha\Delta t & -\frac{\lambda-1+\alpha\Delta t}{2} & 0 & & 0 \\
                    -\frac{\lambda-1+\alpha\Delta t}{2} & \lambda^2-\lambda+\alpha\Delta t & -\frac{\lambda-1+\alpha\Delta t}{2} & \cdots & 0 \\
                    0 & -\frac{\lambda-1+\alpha\Delta t}{2} & \lambda^2-\lambda+\alpha\Delta t & & 0 \\
                    & \vdots & & \ddots & \vdots \\
                    0 & 0 & 0 & \cdots & \lambda^2-\lambda+\alpha\Delta t
                \end{smallmatrix}
            \end{bmatrix}_{n' \times n'}.
        \end{equation*}

    To analyze the eigenvalues of matrix $\bar{A}$ using Gershgorin's disk theorem, 
    we introduce a transformation to equalize the radius of Gershgorin's disk with respect to each eigenvalue.
    Define $P=\diag(p_1,p_2,\ldots,p_{n'})$ with $p_i \in \mathbb R_{\ge 0}$.
    Apparently, $\bar{A}$ is symmetric and $P$ is invertible.
    Therefore, the matrix $P\bar{A}P^{-1}$ presented in \eqref{dadada} (on the top of next page)  and the matrix $A$ have the same eigenvalues.
\begin{figure*}
            \begin{equation}\label{dadada}
            \begin{aligned}
                &P\bar{A}P^{-1} = \\ 
                &\begin{bmatrix}
                        \lambda^2-\lambda+\alpha\Delta t & -\frac{p_2}{2p_1}(\lambda-1+\alpha\Delta t) & 0 & & 0 \\
                        -\frac{p_1}{2p_2}(\lambda-1+\alpha\Delta t) & \lambda^2-\lambda+\alpha\Delta t & -\frac{p_3}{2p_2}(\lambda-1+\alpha\Delta t) & \cdots & 0 \\
                        0 & -\frac{p_2}{2p_3}(\lambda-1+\alpha\Delta t) & \lambda^2-\lambda+\alpha\Delta t & & 0 \\
                        & \vdots & & \ddots & \vdots \\
                        0 & 0 & 0 & \cdots & \lambda^2-\lambda+\alpha\Delta t
                \end{bmatrix}.
            \end{aligned}          
        \end{equation}
\end{figure*}
    According to the Gershgorin's disk theorem,
    each eigenvalue of $\bar{A}$ lies within at least one of the discs centered at $ \lambda^2-\lambda+\alpha\Delta t $ with radiuses
    $\frac{p_2}{2p_1}\cdot(\lambda-1+\alpha\Delta t),\frac{p_1+p_3}{2p_2}\cdot(\lambda-1+\alpha\Delta t),\ldots,\frac{p_{n'-1}}{2p_{n'}}\cdot(\lambda-1+\alpha\Delta t)$, respectively. 
   The constants $p_1, p_2, \cdots, p_{n'}$ are appropriately chosen such that 
    \begin{equation*}\label{3.12}
        \begin{aligned}
            \frac{p_2}{p_1}=\frac{p_1+p_3}{p_2}= \cdots = \frac{p_{i-2}+p_i}{p_{i-1}}= \cdots =\frac{p_{n'-1}}{p_{n'}}.
        \end{aligned}
    \end{equation*}
   By setting $p_1 = 1$, one has 
    \begin{equation}\label{pi:p1}
        \begin{aligned}
            p_i= p_2 p_{i - 1} - p_{i-2}, i\in \{3, 4, \cdots, n'\}.
        \end{aligned}
    \end{equation}
    The solution to \eqref{pi:p1} is given by 
    \begin{equation*}
        \begin{aligned}
            p_i = \frac{\sin(i\arctan\frac{\sqrt{4-p_2^2}}{p_2})}{\sin(\arctan\frac{\sqrt{4-p_2^2}}{p_2})}.
        \end{aligned}
    \end{equation*}
    Recalling that $\frac{p_{n'-1}}{p_n'}=p_2$, there holds
    \begin{equation*}
        \begin{aligned}
            \frac{\sin((n'-1)\arctan\frac{\sqrt{4-p_2^2}}{p_2})}{\sin(n'\arctan\frac{\sqrt{4-p_2^2}}{p_2})}=p_2
        \end{aligned}
    \end{equation*}
  and further $p_2=2\cos(\frac{\pi}{n'+1})$.
    Therefore, all the eigenvalues of $\bar{A} $ lie within the Gerschgorin's disk with 
    $\lambda^2-\lambda+\alpha\Delta t$ being its center and $\left|\cos(\frac{\pi}{n'+1})(\lambda-1+\alpha\Delta t)\right |$
     the radius.
       Since $\lambda $ is the eigenvalue of $A$,
    one has $|\bar{A}| = |\lambda I- A|=0$,
    which means $\bar{A}$ has an eigenvalue $0$.
    In such a case, the origin should be included in the Gerschgorin's disk, 
    which requires $|\lambda^2-\lambda+\delta|\leq |\cos(\frac{\pi}{n'+1})(\lambda-1+\delta)|$.
    
    To compare the radius with the distance between the center and the origin, we  
    define an auxiliary function as
     \begin{equation*}
        \begin{aligned}
            \psi (\lambda) = |\lambda^2-\lambda+\delta|^2-\cos^2(\frac{\pi}{n'+1})|\lambda-1+\delta|^2
        \end{aligned}
    \end{equation*}
    which is the difference between the squared form.
    Next, we will prove that $|\lambda^2-\lambda+\delta|\leq |\cos(\frac{\pi}{n'+1})(\lambda-1+\delta)|$ holds only when $\lambda <1$. Now we give the proof by contradiction, that is, $\psi (\lambda)>0$ always holds if $|\lambda| \geq 1$.

    Assume that the magnitude and the argument of $\lambda$ is $a$ and $\theta$ respectively. Then
    $\lambda$ can be written as  
    \begin{equation}
        \begin{aligned}
            \lambda = a(\sin \theta + \iota  \cos \theta),
        \end{aligned}
    \end{equation}
    where $\iota$ represents the imaginary quantity.
    Accordingly we have 
    \begin{equation}
        \begin{aligned}
            \psi(a) = &a^4-2a^3\sin\theta+2a^2\delta\sin^2\theta-2a^2\delta\cos^2\theta\\
            &-2a\delta\sin\theta+\delta^2+a^2\\
            &+(1-\epsilon)(a^2+2a\sin\theta\delta-2a\sin\theta+\delta^2-2\delta+1).
        \end{aligned}
    \end{equation}
    We now focus on the value of function $\psi(a)$ with respect to $a$.
    If the three inequalities: $\psi (a)|_{a=1}>0$, the first-order partial derivative $\left.\frac{\partial \psi(a)}{\partial a}\right |_{a=1} >0$ 
    and the second-order partial derivative $\left.\frac{\partial^2 \psi(a)}{\partial a^2}\right |_{a\geq 1}>0$ holds,
    one has $\psi (a)>0$ when $a\geq 1$. In the following part, we will respectively consider these three situations. 

    	\begin{enumerate}
    		\item the value of $\psi (a)|_{a=1}$
    		\begin{equation*}
    			\begin{aligned}
    				\psi(a) &= \delta(4\sin^2\theta-4\sin\theta) +\epsilon((2-2\sin\theta)(1-\delta)+\delta^2)\\
    				&>\delta(4\sin^2\theta-4\sin\theta)+\epsilon((2-2\sin\theta)(1-\delta)).
    			\end{aligned}
    		\end{equation*}
    		  Apparently,
    		when $\sin \theta \leq 0$,
    		$\psi(\sin \theta + \iota \cos \theta) > 0$ always holds.
    		when $\sin \theta > 0$,
    		\begin{equation}\label{3.16}
    			\begin{aligned}
    				\psi(a)&>\delta(4\sin\theta-4)+\epsilon((2-2\sin\theta)(1-\delta))\\
    				&>(2-2\sin\theta)(\epsilon(1-\delta)-2\delta).\\
    			\end{aligned}
    		\end{equation}
    		It can be obtained from (\ref{3.16}) that
    		when $\delta < \frac{\epsilon}{2+\epsilon}$,
    		$\psi(\sin \theta + \iota \cos \theta)>0$.
    		
    		\item the value of $\left.\frac{\partial \psi(a)}{\partial a}\right |_{a=1}$
    		
    		The first-order partial derivative of $\psi(a)$ is given by 
    		\begin{equation*}
    			\begin{aligned}
    				\frac{\partial \psi(a)}{\partial a}= &  4a^3-6a^2\sin\theta+4a\delta(\sin^2\theta-\cos^2\theta)-2\delta\sin\theta\\
    				&+2a-(1-\epsilon)(2a+2\sin\theta\delta-2\sin\theta).
    			\end{aligned}
    		\end{equation*}
    		When $a=1$, one has
    		\begin{equation*}\label{3.18}
    			\begin{aligned}
    				\left.\frac{\partial \psi(a)}{\partial a}\right |_{a=1} = &
    				(4-8\delta\sin\theta-4\delta)(1-\sin\theta) \\
    				&+2\epsilon(1+\sin\theta(\delta-1))\\
    				> & (4-12\delta)(1-\sin\theta) \\ 
    				  & +2\epsilon(1+\sin\theta(\delta-1)).\\
    			\end{aligned}
    		\end{equation*}
    		Hence, if $\delta < \frac{1}{3}$, there holds $\left.\frac{\partial \psi(a)}{\partial a}\right |_{a=1} >0$.
    		
    		\item the value of  $\left.\frac{\partial^2 \psi(a)}{\partial a^2}\right |_{a\geq 1}$
    		
    		 Under the condition that $a\geq 1$, the second-order partial derivative of $\psi(a)$ satisfies 
    		\begin{equation*}\label{3.19}
    			\begin{aligned}
    				\frac{\partial^2 \psi(a)}{\partial a^2} = &12a^2-12a\sin\theta+4\delta\sin^2\theta-4\delta\cos^2\theta\\ 
    				&+2-2(1-\epsilon)\\
    				\geq & 12-12\sin\theta+4\delta(\sin^2\theta-1)+2\epsilon\\
    				= & 4(3-\delta\sin\theta-\delta)(1-\sin\theta)+2\epsilon\\
    				 > & 4(3-2\delta)(1-\sin\theta)+2\epsilon.
    			\end{aligned}
    		\end{equation*}
    		It thus can be inferred from the above equation that $\frac{\partial^2 \psi(a)}{\partial a^2}>0$
    		if $\delta<\frac{3}{2}$.
    		
    	\end{enumerate}
      
    To sum up, when $a = |\lambda| \geq 1$ and $\delta< \frac{\epsilon}{2+\epsilon}$,
    the inequality $\psi (a)>0$ always holds,
    which leads to the contradiction.
    Hence, the spectral radius of matrix $A$ is less than 1.

\subsection{Proof of the invertibility}\label{invertibility}
The determinant of $M(d)$ is given by
$$|M(d)| = (1+\beta)M(d-1) - \frac{(1-\beta)^2}{4}M(d-2).$$
By considering $|M(0)| = 1,|M(1)| = 1+\beta$, one has 
\begin{equation*}\label{3.27}
	\begin{aligned}
		|M(d)| =& \underbrace{\frac{2\sqrt{\beta} + \beta + 1}{4\sqrt{\beta}}(\frac{1+\beta}{2} + \sqrt{\beta})^d}_{\dfb M_1}\\
		 &+ \underbrace{\frac{2 \sqrt{\beta} - \beta - 1}{4\sqrt{\beta}}(\frac{1+\beta}{2} - \sqrt{\beta})^d}_{\dfb M_2}.
	\end{aligned}
\end{equation*}
Obviously,
$|M_1|>|M_2|$ always holds when $d\geqslant 0$, implying $|M(d)|\neq 0, \forall d \in \mathbb{N}$.
Therefore, the invertibility of matrix $(I+A_{22}-\beta A_{21})$ is proved.

\subsection{Proof of Theorem 3}\label{app2}
    The characteristic polynomial of $A_r$ is
    \begin{equation}
		\begin{aligned}
			|\lambda I - A_r| &= \left|\begin{array}{ccc}
				(\lambda - 1)I & 0 & -\Delta t * I \\
				0 & \lambda I & -I \\
				-\alpha A_{21} & -A_{22} & \lambda I
			\end{array}\right|\\
		&=\left| (\lambda^3-\lambda^2)I - \lambda\Delta t\alpha A_{21}-(\lambda - 1)A_{22}\right|.
		\end{aligned}
	\end{equation}
    Denote $\bar{A}_r = (\lambda^3-\lambda^2)I - \lambda\Delta t\alpha A_{21}-(\lambda - 1)A_{22}$.
    Similar to the proof scheme of Theorem 1 in Appendix \ref{app1}, it can be shown that
    all the eigenvalues of $\bar{A}_r$ lie within the disk of radius $\cos^2(\frac{\pi}{n'+1})|\lambda-1+\lambda\alpha\Delta t|$ centered at $\lambda^3-\lambda^2+\lambda\alpha\Delta t$. 
    Next, we will prove by contradiction that $|\lambda^3-\lambda^2+\lambda\alpha\Delta t|> \cos^2(\frac{\pi}{n'+1})|\lambda-1+\lambda\alpha\Delta t|$ holds only when $|\lambda|\geq 1$.
    Define an auxiliary function as follows 
    \begin{equation}
        \begin{aligned}
            \psi_r(\lambda) = & |\lambda^2-\lambda+\alpha\Delta t|^2-\cos^2(\frac{\pi}{n+1})|\lambda-1+\lambda\alpha\Delta t|^2\\
            =& |\lambda^2-\lambda+\delta|^2-(1-\epsilon)|\lambda-1+\lambda\delta|^2\\
            =&a^4-2a^3\sin\theta+2a^2\delta\sin^2\theta-2a^2\delta\cos^2\theta\\
            &-2a\delta\sin\theta+\delta^2+a^2\\
            &-(1-\epsilon)\left(a^2(1+\delta)^2-2a\sin\theta\delta-2a\sin\theta+1\right)
        \end{aligned}
    \end{equation}
    Again following the same lines of Appendix \ref{app1}, we consider the following three situations.
        	\begin{enumerate}
    	\item the value of $\psi_r(a)|_{a=1}$
    	
    	In such a situation, we have 
    	\begin{equation*}\label{3.39}
    		\begin{aligned}
    			\psi_r(\lambda) =& -4 \cos^2 \theta + \epsilon(\delta^2+2\delta + 2-2\sin \theta - 2\delta\sin \theta)\\
    			>&-2\delta(2-2\sin\theta)(2+2\sin\theta) \\
    			& + \epsilon(\delta + 1)(2-2\sin\theta)\\
    			\geq & (2-\sin\theta) (\epsilon \delta + \epsilon - 6\delta)
    		\end{aligned}
    	\end{equation*}
    	It can be derived that $\psi_r(\lambda) > 0$ holds when $\delta < \frac{\epsilon}{6 - \epsilon}$.

    	\item the value of $\left.\frac{\partial \psi_r(a)}{\partial a}\right |_{a=1}$
    	
    	  Taking the derivative of $\psi_r$ w.r.t. a gives   
    	   \begin{equation*}
    		\begin{aligned}
    			\frac{\partial \psi_r(\lambda)}{\partial a} =& 4a^3-6a^2\sin\theta+4a\delta\sin^2\theta-4a\delta\cos^2\theta \\
    			 &-(1-\epsilon)\left(2a(1+\delta)^2-2\sin\theta\delta-2\sin\theta\right)\\ 
    			 & -2\delta\sin\theta+2a.
    		\end{aligned}
    	\end{equation*}
    	By letting $a=1$, there holds 
    	\begin{equation*}
    		\begin{aligned}
    			\left.\frac{\partial \psi_r(\lambda)}{\partial a}\right |_{a=1} =&4-4\sin\theta + 4\delta\sin^2\theta-4\delta\cos^2\theta-2\delta^2 \\
    			 & -4\delta+\epsilon\left(2(\delta+1)^2-2\sin\theta(\delta+1)\right)\\
    			=&(4-8\delta - 8\delta\sin\theta)(1-\sin\theta) + \epsilon(2(\delta+1)^2\\
    			&-2\sin\theta(\delta+1)) - 2\delta^2\\
    			>&(4-12\delta)(1-\sin\theta) + 2\delta(\epsilon - \delta) + 2\epsilon\delta^2.
    		\end{aligned}
    	\end{equation*}
    	Hence, $\left.\frac{\partial \psi_r(\lambda)}{\partial a}\right |_{a=1} >0$ holds if $\delta < \min(\frac{1}{3},\epsilon)$.

       	\item the value of  $\left.\frac{\partial^2 \psi_r(a)}{\partial a^2}\right |_{a\geq 1}$ 	
       	
       	By direct calculation, one has 
       	 \begin{equation*}
       		\begin{aligned}
       			\frac{\partial^2 \psi_r(\lambda)}{\partial a^2}=&12a^2-12a\sin\theta+4\delta\sin^2\theta-4\delta\cos^2\theta+2\\
       			&-2(1-\epsilon)(1+\delta)^2\\
       			>&4(3-2\delta)(1-\sin\theta)+2-2(1-\epsilon)(1+\delta)^2.
       		\end{aligned}
       	\end{equation*}
    	Accordingly, when $\delta < \min(\frac{3}{2},\frac{\epsilon}{1-\epsilon})$, the second-order derivative of $\psi_r$ is always positive.
    	    	
    	\end{enumerate}
            
    To sum up, $\psi_r(\lambda)>0$, when $\delta < \frac{\epsilon}{6-\epsilon}, \forall a\geq1$,
    which leads to the contradiction. Therefore, the spectral radius of $A_r$ is less than $1$.

\bibliographystyle{Bibliography/IEEEtranTIE}
\bibliography{Bibliography/IEEEabrv}\ 

\begin{IEEEbiography}[{\includegraphics[width=1in,height=1.25in,clip,keepaspectratio]{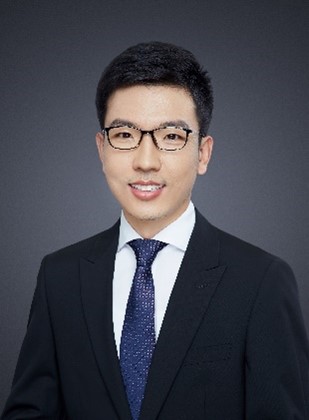}}]
	{Qingkai Yang}(Member, IEEE) received the first Ph.D. degree in control science and engineering from the Beijing Institute of Technology, Beijing, China, in 2018, and the second Ph.D. degree in system control from the University of Groningen, Groningen, The Netherlands, in 2018. 

	He is currently an Associate Professor with the School of Automation, Beijing Institute of Technology. His research interest is in formation control of multi-agent systems and intelligent robotics.
	\end{IEEEbiography}
	
	\begin{IEEEbiography}[{\includegraphics[width=1in,height=1.25in,clip,keepaspectratio]{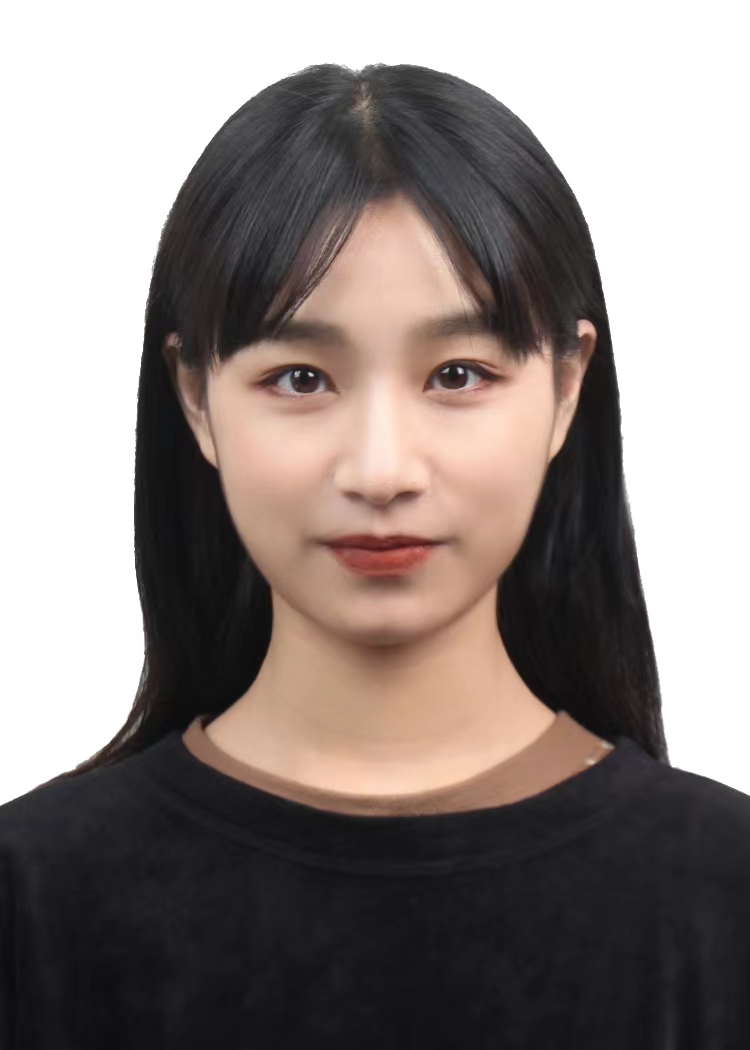}}]
	{Fan Xiao} received the B.S. degree in automation from the Beijing Institute of Technology, Beijing, China, in 2020,
	and is currently pursuing the M.S. degree in control science and engineering from the Beijing Institute of Technology, Beijing, China.

	Her research interests include control of multi-agent systems and topology optimization.
	\end{IEEEbiography}
	
	\begin{IEEEbiography}[{\includegraphics[width=1in,height=1.25in,clip,keepaspectratio]{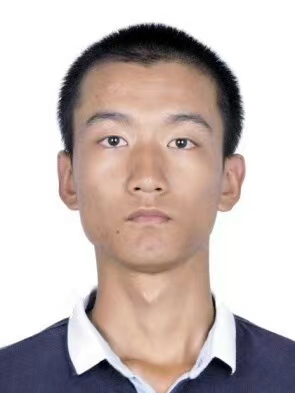}}]
		{Jingshuo Lyv} Jingshuo Lyv received the B.S. degree in automation from the Beijing Institute of Technology, Beijing, China, in 2021, and is currently pursuing the M.S. degree in control science and engineering from the Beijing Institute of Technology, Beijing, China.
		
		His research interests include data-driven control and wind disturbance rejection.
	\end{IEEEbiography}
	
	\begin{IEEEbiography}[{\includegraphics[width=1in,height=1.25in,clip,keepaspectratio]{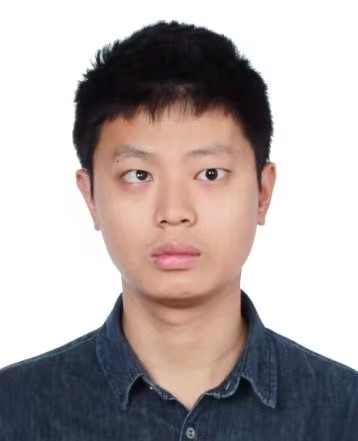}}]
		{Bo Zhou} received the M.S. degree in control science and engineering from the Beijing Institute of Technology, Beijing, China, in 2021.
		
		His research interests include formation control and distributed estimation.
	\end{IEEEbiography}

	\begin{IEEEbiography}[{\includegraphics[width=1in,height=1.25in,clip,keepaspectratio]{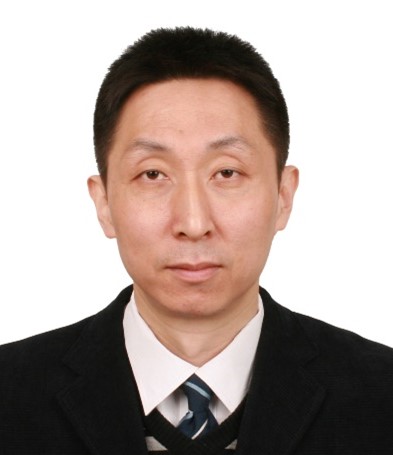}}]
	{Hao Fang}(Member, IEEE) received the B.S. degree from the Xi’an University of Technology, Shaanxi, China, in 1995, and the M.S. and Ph.D. degrees from the Xi’an Jiaotong University, Shaanxi, in 1998 and 2002, respectively.

	Since 2011, he has been a Professor with the Beijing Institute of Technology, Beijing, China. 
	He held two postdoctoral appointments with the INRIA/France Research Group of COPRIN and the LASMEA (UNR6602 CNRS/Blaise Pascal University, Clermont-Ferrand, France). 
	His research interests include all-terrain mobile robots, robotic control, and multiagent systems.

	\end{IEEEbiography}

\end{document}